\documentclass{article}

\usepackage{amsmath,amsbsy,amsfonts,amssymb,dsfont,units,psfrag}%amsthm,

\usepackage{algorithmic}
\usepackage{algorithm,mathtools}
\usepackage{enumitem}
\usepackage{color,cases,caption}

\usepackage{tikz,lipsum}
\usetikzlibrary{calc, shapes, backgrounds, arrows, fit, calc, positioning}
\usepackage{hyperref}
\usepackage[numbers]{natbib}
 \newcommand{\IGNORE}[1]{}

\usepackage{subcaption}
\usepackage{float}
\makeatletter
\renewcommand*{\@opargbegintheorem}[3]{\trivlist
      \item[\hskip \labelsep{\bfseries #1\ #2}] \textbf{(#3)}\ \itshape}
\makeatother
\newcommand\R{\mathbb{R}}

\def\Pc{{\cal S}}

\makeatletter
\def\thm@space@setup{%
  \thm@preskip=\parskip \thm@postskip=0pt
}
\makeatother

\DeclareMathOperator{\poly}{poly}

\newcommand\tl{\tilde}

\usepackage{fullpage}

\def\tl{\tilde}

\newcommand\inner[1]{\ensuremath{\langle #1 \rangle}}

\def\tha{{\mbox{\tiny th}}}

\DeclareMathOperator{\Diag}{Diag}

 \def\0{{\bf 0}}

\def\viz{{viz.,\ \/}}

\def\qed{\hfill\hbox{${\vcenter{\vbox{
    \hrule height 0.4pt\hbox{\vrule width 0.4pt height 6pt
    \kern5pt\vrule width 0.4pt}\hrule height 0.4pt}}}$}}

\definecolor{myred}{rgb}{0.3,0.0,0.7}
\definecolor{dkg}{rgb}{0.1,0.7,0.2}
\definecolor{dkb}{rgb}{0.0,0.2,0.8}
\definecolor{brm}{rgb}{1,0.0,1}

\def\Sc{{\cal S}}

\def\Ebb{{\mathbb E}}

\def\Rbb{{\mathbb R}}

\newcommand{\bprf}{\begin{myproof}}
\newcommand{\eprf}{\end{myproof}}
\newcommand{\bp}{\begin{psfrags}}
\newcommand{\ep}{\end{psfrags}}
\newcommand{\bl}{\begin{lemma}}
\newcommand{\el}{\end{lemma}}
\newcommand{\bt}{\begin{theorem}}
\newcommand{\et}{\end{theorem}}
\newcommand{\bc}{\begin{center}}
\newcommand{\ec}{\end{center}}
\newcommand{\bi}{\begin{itemize}}
\newcommand{\ei}{\end{itemize}}
\newcommand{\ben}{\begin{enumerate}}
\newcommand{\een}{\end{enumerate}}
\newcommand{\bd}{\begin{definition}}
\newcommand{\ed}{\end{definition}}
\def\beq{\begin{equation}}
\def\eeq{\end{equation}\noindent}
\def\beqn{\begin{eqnarray}}
\def\eeqn{\end{eqnarray} \noindent}
\def\beqnn{  \begin{eqnarray*}}
\def\eeqnn{\end{eqnarray*}  \noindent}
\def\bcase{  \begin{numcases}}
\def\ecase{\end{numcases}   \noindent}
\def\bsbcase{  \begin{subnumcases}}
\def\esbcase{\end{subnumcases}   \noindent}

\newenvironment{myproof}{\noindent{\bf Proof:} \hspace*{1em}}{
    \hspace*{\fill} $\Box$ }
\newenvironment{proof_of}[1]{\noindent {\bf Proof of #1: }}{\hspace*{\fill} $\Box$ }

\newcommand{\matplottc}[1]{               % single matlab plot twocolumn
        \unitlength .45truein
        \begin{center}
        \includegraphics{#1.ps}
        \end{picture}
        \end{center}
}

\def\psfancypar#1#2{\begingroup\def\par{\endgraf\endgroup\lineskiplimit=0pt}
               \setbox2=\hbox{\large\sc #2}
               \newdimen\tmpht \tmpht \ht2 \advance\tmpht by \baselineskip
               \font\hhuge=Times-Bold at \tmpht
               \setbox1=\hbox{{\hhuge #1}}
               \count7=\tmpht \count8=\ht1
               \divide\count8 by 1000 \divide\count7 by \count8
               \tmpht=.001\tmpht\multiply\tmpht by \count7
               \font\hhuge=Times-Bold at \tmpht
               \setbox1=\hbox{{\hhuge #1}}
               \noindent
                \hangindent1.05\wd1
               \hangafter=-2 {\hskip-\hangindent
               \lower1\ht1\hbox{\raise1.0\ht2\copy1}%
                \kern-0\wd1}\copy2\lineskiplimit=-1000pt}

\def\Kout{\setbox1=\hbox{\Huge\bf K}\hbox to
1.05\wd1{\hspace{.05\wd1}
}}
\def\Sout{\setbox1=\hbox{\Huge\bf S}\hbox to 1.05\wd1{\hspace{.05\wd1}
}}

\newcommand{\torestate}[3]{%
\expandafter \def \csname BBRESTATE #2 \endcsname{#3}
\theoremstyle{plain}
\newtheorem{BBRESTATETHMNUM#2}[theorem]{#1}
\begin{BBRESTATETHMNUM#2}\label{#2}\csname BBRESTATE #2 \endcsname   \end{BBRESTATETHMNUM#2}
\newtheorem*{BBRESTATETHMNONNUM#2}{{#1}~\ref{#2}}
}

\newcommand{\restate}[1]{\begin{BBRESTATETHMNONNUM#1}[Restated] \csname BBRESTATE #1 \endcsname
\end{BBRESTATETHMNONNUM#1}}

\definecolor{blue1}{HTML}{0066FF}
\definecolor{lpurple}{cmyk}{.05,0.18,0,0}
\newtheorem{definition}{Definition}

\newtheorem{theorem}{Theorem}
\newtheorem{lemma}[theorem]{Lemma}

    \newtheorem{proof}{Proof}

\begin{document}

\title{Training Input-Output Recurrent Neural Networks\\ through Spectral Methods}
 % Authors with different addresses:
%\usepackage{fullpage}

\author{Hanie Sedghi\footnote{Allen institute For Artificial Intelligence. Email: hanies@allenai.org} \quad Anima Anandkumar\footnote{University of California, Irvine. Email: a.anandkumar@uci.edu}}

\maketitle

\begin{abstract}We consider the problem of training input-output recurrent neural networks (RNN) for sequence labeling tasks. We propose a novel spectral approach for learning the network parameters. It is based on  decomposition of the cross-moment tensor between the output and a non-linear transformation of the input, based on score functions. We guarantee consistent learning with polynomial sample and computational complexity under transparent conditions such as   non-degeneracy of model parameters, polynomial activations  for the neurons, and a Markovian evolution of the input sequence. We also extend our results to   Bidirectional RNN which uses both previous and  future information to output the label at each time point, and is employed in many NLP tasks such as POS tagging.
 \end{abstract}


\section{Introduction}

\begin{figure}[t]
\hspace{-4em}
\subcaptionbox{\small NN}{
\begin{minipage}[t]{0.3\textwidth}
\begin{center}
\resizebox{.7\textwidth}{!}{
\begin{tikzpicture}
  [
    scale=1,
    observed/.style={circle,minimum size={width("$x_{d}$")+15pt},inner
sep=0mm,draw=violet,fill=lpurple,line width=.5mm},
    hidden/.style={circle,minimum size={width("$x_{d}$")+15pt},inner sep=0mm,draw=dkg,line width=.5mm},
        func/.style={circle,minimum size=0.6cm,inner sep=0mm,draw=blue1,dashed, line width=.5mm},
        vdots/.style={min, node distance=.5mm},
  ]
  \tikzset{
    %Define standard arrow tip
    >=stealth',
    pil/.style={
           ->,
           thick,
           shorten <=2pt,
           shorten >=2pt,}
}
  \node [hidden,name=f1] at ($(-2.5,0)$) {};
  \node [hidden,name=f2] at ($(-.5,0)$) {};

  \node[observed,name=y1] at ($(-1.5,2)$){};%{$y_1$};

 \node [observed,name=x1] at ($(-1.5,-2)$) {}; %$x_1$

  \node [] at ($(-3,-2)$) {Input};
  \node [] at ($(-3,2)$) {Output};
    \node [] at ($(-4.2,0)$) {Hidden Layer};

  \draw [blue, line width=.5mm, ->] (f1) to (y1);
  \draw [blue, line width=.5mm, ->] (f2) to (y1);

   \draw [red, line width=.5mm, <-] (f1) to (x1);
  \draw [red, line width=.5mm, <-] (f2) to (x1);
    \node [] at ($(-1.5,-3)$) {any $t$};

\end{tikzpicture}
}
\end{center}
\end{minipage}
}
\hspace{-2.5em}
\subcaptionbox{\small  IO-RNN}{
\begin{minipage}[t]{0.3\textwidth}
\begin{center}
\resizebox{1.2\textwidth}{!}{
\begin{tikzpicture}
  [
    scale=1,
    observed/.style={circle,minimum size={width("$x_{d}$")+15pt},inner
sep=0mm,draw=violet,fill=lpurple,line width=.5mm},
    hidden/.style={circle,minimum size={width("$x_{d}$")+15pt},inner sep=0mm,draw=dkg,line width=.5mm},
        func/.style={circle,minimum size=0.6cm,inner sep=0mm,draw=blue1,dashed, line width=.5mm},
        vdots/.style={min, node distance=.5mm},
  ]
  \tikzset{
    %Define standard arrow tip
    >=stealth',
    pil/.style={
           ->,
           thick,
           shorten <=2pt,
           shorten >=2pt,}
}
  \node [hidden,name=f1] at ($(-2.5,0)$) {};
  \node [hidden,name=f2] at ($(-.5,0)$) {};
  \node [hidden,name=fk] at ($(1.5,0)$) {};
 \node [hidden,name=fn] at ($(3.5,0)$) {};
%  \node [] at ($(-2.1,0.6)$) {$1$};
%  \node [] at ($(2.1,0.6)$) {$k$};
%    \node [hidden,name=h1] at ($(-2,1)$){};% {$h_{11}$};
%  \node [hidden,name=h2] at ($(-1,1)$){};% {$h_{12}$};
%  \node [hidden,name=hn] at ($(1,1)$){};
%  \node [hidden,name=hk] at ($(2,1)$){};% {$h_{1k}$};
%  \node [] at ($(-2.7,1)$) {$\tcdkg{h}$};
  \node[observed,name=y1] at ($(-1.5,2)$){};%{$y_1$};
  %  \node [observed,name=y] at ($(0,-4.5)$) {$y$};
  \node [observed,name=y2] at ($(2.5,2)$){};%{$y_{d_y}$};
%  \node [observed,name=yd] at ($(1.5,-3.5)$) {$y_{d_y}$};
 \node [observed,name=x1] at ($(-1.5,-2)$) {}; %$x_1$
%   \node [observed,name=x1] at ($(-1.5,-2)$) {$x_2$}; %$x_2$
  \node [observed,name=x2] at ($(2.5,-2)$) {}; %$x_3$
 % \node [observed,name=xd1] at ($(1.5,1.5)$) {}; %$x_{n_x-1}$
%  \node [observed,name=xd2] at ($(2.5,-2)$) {$x_{d}$}; %$x_{n_x}$
  %\node [] at ($(-2.2,-4.5)$) {$y$};
  \node [] at ($(-3,-2)$) {Input};
  \node [] at ($(-3,2)$) {Output};
    \node [] at ($(-4.2,0)$) {Hidden Layer};
        \node [] at ($(2.5,-3)$) {$t=2$};
         \node [] at ($(-1.5,-3)$) {$t=1$};
%  \node [] at ($(-2.25,1.2)$) {\tcb{$A_2$}};
%   \node [] at ($(-2.7,-1)$) {\tcr{$A_1$}};
%  \node at ($(1,-1)$) {$\dotsb$};
%    \node at ($(0.5,2)$) {$\dotsb$};
%        \node at ($(1,-1.2)$) {$\dotsb$};
%  \node at ($(0,0)$) {$\dotsb$};
%   \node at ($(1,-2)$) {$\dotsb$};

  \draw [blue, line width=.5mm, ->] (f1) to (y1);
  \draw [blue, line width=.5mm, ->] (f2) to (y1);
%  \draw [blue, line width=.5mm, ->] (f1) to (yk);
%% \draw [red, line width=.5mm, ->] (h22) to (g3);
% \draw [blue, line width=.5mm,->] (f2) to (y2);
%  \draw [blue, line width=.5mm, ->] (f2) to (y1);
  \draw [blue, line width=.5mm, ->] (fn) to (y2);
  \draw [blue, line width=.5mm, ->] (fk) to (y2);
% % \draw [red, line width=.5mm, ->] (h2k) to (g1);
%   \draw [blue, line width=.5mm, ->] (fn) to (y2);
%   \draw [blue, line width=.5mm, ->] (fn) to (yk);
%
   \draw [red, line width=.5mm, <-] (f1) to (x1);
  \draw [red, line width=.5mm, <-] (f2) to (x1);
%   \draw [red, line width=.5mm, <-] (f2) to (x11);
  \draw [red, line width=.5mm,<-] (fn) to (x2);
  \draw [red, line width=.5mm,<-] (fk) to (x2);
%  \draw [red, line width=.5mm, <-] (fk) to (x1);
%  \draw [red, line width=.5mm, <-] (fn) to (x2);
%
%  \draw [red, line width=.5mm,<-] (fn) to (xd2);
%\draw [gray] (6,0) arc [dashed, thick, radius=1, start angle=45, end angle= 120];
\draw[dashed, very thick, ->] (f1) to [out=30,in=130] (fk);
\draw[dashed, very thick, ->] (f1) to [out=45,in=130] (fn);
\draw[dashed, very thick, ->] (f2) to [out=330,in=230] (fk);
\draw[dashed, very thick, ->] (f2) to [out=315,in=230] (fn);

\end{tikzpicture}
}
\end{center}
\end{minipage}
}
\hspace{1em}
\subcaptionbox{\small  BRNN}{
\begin{minipage}[t]{0.38\textwidth}
\begin{center}
\resizebox{1.3\textwidth}{!}{
\begin{tikzpicture}
  [
    scale=1,
    observed/.style={circle,minimum size={width("$x_{d}$")+15pt},inner
sep=0mm,draw=violet,fill=lpurple,line width=.5mm},
    hidden/.style={circle,minimum size={width("$x_{d}$")+15pt},inner sep=0mm,draw=dkg,line width=.5mm},
        func/.style={circle,minimum size=0.6cm,inner sep=0mm,draw=blue1,dashed, line width=.5mm},
        vdots/.style={min, node distance=.5mm},
  ]
  \tikzset{
    %Define standard arrow tip
    >=stealth',
    pil/.style={
           ->,
           thick,
           shorten <=2pt,
           shorten >=2pt,}
}
  \node [hidden,name=f1] at ($(-2.5,-.8)$) {};
  \node [hidden,name=f2] at ($(-.5,0)$) {};
  \node [hidden,name=f3] at ($(1.5,-.8)$) {};
 \node [hidden,name=f4] at ($(3.5,0)$) {};
   \node [hidden,name=f5] at ($(5.5,-.8)$) {};
 \node [hidden,name=f6] at ($(7.5,0)$) {};
%  \node [] at ($(-2.1,0.6)$) {$1$};
%  \node [] at ($(2.1,0.6)$) {$k$};
%    \node [hidden,name=h1] at ($(-2,1)$){};% {$h_{11}$};
%  \node [hidden,name=h2] at ($(-1,1)$){};% {$h_{12}$};
%  \node [hidden,name=hn] at ($(1,1)$){};
%  \node [hidden,name=hk] at ($(2,1)$){};% {$h_{1k}$};
%  \node [] at ($(-2.7,1)$) {$\tcdkg{h}$};
  \node[observed,name=y1] at ($(-1.5,2)$){};%{$y_1$};
  %  \node [observed,name=y] at ($(0,-4.5)$) {$y$};
  \node [observed,name=y2] at ($(2.5,2)$){};%{$y_{d_y}$};
    \node [observed,name=y3] at ($(6.5,2)$){};
%  \node [observed,name=yd] at ($(1.5,-3.5)$) {$y_{d_y}$};
 \node [observed,name=x1] at ($(-1.5,-2.5)$) {}; %$x_1$
%   \node [observed,name=x1] at ($(-1.5,-2)$) {$x_2$}; %$x_2$
  \node [observed,name=x2] at ($(2.5,-2.5)$) {}; %$x_3$
    \node [observed,name=x3] at ($(6.5,-2.5)$) {};
 % \node [observed,name=xd1] at ($(1.5,1.5)$) {}; %$x_{n_x-1}$
%  \node [observed,name=xd2] at ($(2.5,-2)$) {$x_{d}$}; %$x_{n_x}$
  %\node [] at ($(-2.2,-4.5)$) {$y$};
  \node [] at ($(-3,-2)$) {Input};
  \node [] at ($(-3,2)$) {Output};
    \node [] at ($(-4.2,0)$) {Hidden Layer};
        \node [] at ($(2.5,-3.3)$) {$x_t$};
         \node [] at ($(-1.5,-3.3)$) {$x_{t-1}$};
          \node [] at ($(6.5,-3.3)$) {$x_{t+1}$};
             \node [] at ($(2.5,2.8)$) {$y_t$};
         \node [] at ($(-1.5,2.8)$) {$y_{t-1}$};
          \node [] at ($(6.5,2.8)$) {$y_{t+1}$};
            \node [] at ($(-2.7,-1.4)$) {$z_{t-1}$};
  \node [hidden,name=f2] at ($(-.5,0)$) {};
  \node [hidden,name=f3] at ($(1.5,-.8)$) {};
 \node [hidden,name=f4] at ($(3.5,0)$) {};
   \node [hidden,name=f5] at ($(5.5,-.8)$) {};
 \node [hidden,name=f6] at ($(7.5,0)$) {};
%  \node [] at ($(-2.25,1.2)$) {\tcb{$A_2$}};
%   \node [] at ($(-2.7,-1)$) {\tcr{$A_1$}};
%  \node at ($(1,-1)$) {$\dotsb$};
%    \node at ($(0.5,2)$) {$\dotsb$};
%        \node at ($(1,-1.2)$) {$\dotsb$};
%  \node at ($(0,0)$) {$\dotsb$};
%   \node at ($(1,-2)$) {$\dotsb$};

  \draw [blue, line width=.5mm, ->] (f1) to (y1);
  \draw [orange, line width=.5mm, ->] (f2) to (y1);
%  \draw [blue, line width=.5mm, ->] (f1) to (yk);
%% \draw [red, line width=.5mm, ->] (h22) to (g3);
% \draw [blue, line width=.5mm,->] (f2) to (y2);
%  \draw [blue, line width=.5mm, ->] (f2) to (y1);
  \draw [blue, line width=.5mm, ->] (f3) to (y2);
  \draw [orange, line width=.5mm, ->] (f4) to (y2);
    \draw [blue, line width=.5mm, ->] (f5) to (y3);
  \draw [orange, line width=.5mm, ->] (f6) to (y3);
% % \draw [red, line width=.5mm, ->] (h2k) to (g1);
%   \draw [blue, line width=.5mm, ->] (fn) to (y2);
%   \draw [blue, line width=.5mm, ->] (fn) to (yk);
%
   \draw [red, line width=.5mm, <-] (f1) to (x1);
  \draw [teal, line width=.5mm, <-] (f2) to (x1);
%   \draw [red, line width=.5mm, <-] (f2) to (x11);
  \draw [red, line width=.5mm,<-] (f3) to (x2);
  \draw [teal, line width=.5mm,<-] (f4) to (x2);
  \draw [red, line width=.5mm, <-] (f5) to (x3);
  \draw [teal, line width=.5mm, <-] (f6) to (x3);
%
%  \draw [red, line width=.5mm,<-] (fn) to (xd2);
%\draw [gray] (6,0) arc [dashed, thick, radius=1, start angle=45, end angle= 120];
\draw[dashed, very thick, ->] (f2) to  (f4);
\draw[dashed, very thick, ->] (f4) to  (f6);
\draw[olive][ dashed, very thick, <-] (f1) to  (f3);
\draw[olive][ dashed, very thick, <-] (f3) to  (f5);
%\draw[dashed, very thick, ->] (fk) to  (f1);
%\draw[dashed, very thick, ->] (fn) to  (f1);
%\draw[dashed, very thick, ->] (fk) to (f2);
%\draw[dashed, very thick, ->] (fn) to  (f2);

\end{tikzpicture}

}
\end{center}
\end{minipage}
}
\caption{\small Graphical representation of a Neural Network (NN) versus an Input-Output Recurrent Neural Network (IO-RNN) and a Bidirectional Recurrent Neural Network (BRNN)}
\label{fig:RNN}
\end{figure}
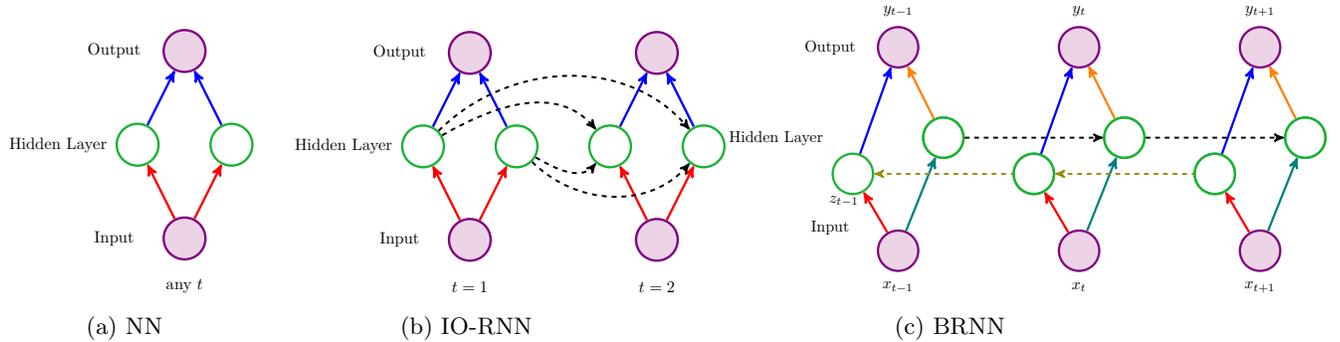

Learning with sequential data is widely encountered in   domains such as natural language processing, genomics, speech recognition, video processing, financial time series analysis,  and so on. Recurrent neural networks (RNN) are a flexible class of sequential  models which can memorize   past information, and selectively pass it on across sequence steps on multiple scales. However, training  RNNs is   challenging in practice, and backpropagation  suffers from exploding and vanishing gradients as the length of the training sequence grows. To overcome this, either RNNs are trained over short sequences or incorporate more complex architectures such as long short-term memories (LSTM). For a detailed overview of RNNs, see~\citep{lipton2015critical}.  Figure~\ref{fig:RNN} contrasts the RNN with a feedforward neural network which has no memory.

On the theoretical front, 
%our 
understanding of RNNs is at best rudimentary. With the current techniques, it is not tractable to  analyze the highly non-linear state evolution in RNNs. Analysis of  backpropagation is also intractable due to non-convexity of the loss function, and in general, reaching the global optimum is hard.  Here, we take the first steps towards addressing these challenging issues. We design novel spectral methods for training IO-RNN and BRNN models.
% in this paper.

We consider the class of input-output RNN or IO-RNN models, where each input in the sequence $x_t$ has an output label $y_t$.  These are useful for  sequence labeling tasks,  which  has many applications such as parts of speech (POS) tagging and named-entity recognition (NER) in NLP~\citep{manning1999foundations},  motif finding in protein analysis~\citep{ben2006sequence}, action recognition in videos~\citep{karpathy2014large}, and so on.

In addition, we also consider an extension of IO-RNN, \viz  the bi-directional RNN or  BRNN, first proposed by~\citet{schuster1997bidirectional}. This  
%FORLATERUSE incorporates information from both sides of the input sequence for hidden state evolution, i.e.  it 
includes two classes of hidden neurons: the first class receives recurrent connection from previous states, and the second class receives it from next steps. See Figure~\ref{fig:RNN}(c). BRNN is useful in NLP tasks such as POS tagging, where both previous and next words in a sentence have an effect on labeling the current word.

In this paper, we develop novel spectral methods for training IO-RNN and BRNN models. Spectral methods have previously been employed for unsupervised learning of a range of latent variable models such as hidden Markov models (HMM), topic models, network community models, and so on~\citep{JMLR:v15:anandkumar14b}. The idea is to decompose moment matrices and tensors using computationally efficient algorithms. The recovered components of the tensor decomposition yield consistent estimates of the model parameters. However, a direct application of these techniques is ruled out due to  non-linearity of the activations in the RNN.

Recently,~\citet{janzamin2014matrix} derived a new framework  for training input-output models in the supervised framework. It is based on spectral decomposition of moment tensors, obtained after certain non-linear transformations of the input. These non-linear transformations take the form of  {\em score functions},  which depend on   generative models of the input. This   provides a new approach for  transferring generative information, obtained through unsupervised learning, into   discriminative training on labeled samples.
Based on the aforementioned approach,~\citet{JanzaminEtal:NN2015}  provided guaranteed risk bounds for training two-layer feedforward neural network models  with polynomial sample and computational bounds. The conditions for obtaining the risk bounds are mild: a small approximation error for the target function under the given class of neural networks, a generative input model with a continuous distribution, and general sigmoidal activations at the neurons.

We propose new spectral approaches for training IO-RNNs in both classification and regression settings. The previous score function approach for training feedforward networks (as described above) does not immediately extend, and there are some non-trivial challenges: (i) Non-linearity in a RNN is propagated along multiple steps in the sequence, while in the two-layer feedforward network, non-linearity is applied only once to the input. It is not immediately clear on how to ``untangle'' all these non-linearities and obtain guaranteed estimates of the network weights. (ii) Learning   bidirectional RNNs is even more challenging since recursive non-linearities are applied in both the directions, and (iii) Assumption of  i.i.d. input and output training samples is no longer applicable, and analyzing concentration bounds for samples generated from a  RNN with non-linear state evolution is challenging. We address all these challenges concretely in this paper.

\subsection{Summary of Results}

%\hscomment{this is regression setting not classification}

%\hscomment{some notes on why polynomial. It has been used before, for example~\citet{xie2015recurrent} uses polynomial RNN for dialogue state tracking in semantic parsers.}

%\hscomment{we need to say what are the cases where our assumptions are true and why it is ok to make such assumptions} 

%\hscomment{Solving for RNN is highly non convex and we have made significant progress in overcoming this. There is no reason to only use ReLUs. Polynomial activations provide another means for non-linearity, and low degree polynomials provide good representations. In fact, even for input-output HMMs there was previously no known technique (emphasize on this more)}

%\hscomment{We are assuming a geometrically ergodic Markov chain not any Markov chain}
Our main contributions are: (i) novel approaches for training input-output RNN and bidirectional RNN models using tensor decomposition methods, (ii) guaranteed recovery of network parameters with polynomial computational and sample complexity, and (iii) transparent conditions for successful recovery based on non-degeneracy of model parameters and bounded evolution of hidden states.

\textbf{Score function transformations: }Training input-output neural networks under arbitrary input is computationally hard. On the other hand, we show that training becomes tractable through spectral methods when the input is generated from a probabilistic model on a continuous state space. This paper can be considered as study of what it takes to uncover the nonlinear dynamics in the system. Since learning under arbitrary input is extremely challenging, we seek to discover under what functions/information of the input the problem becomes tractable. Although this differs from usual approach in training IO-RNNs, this is a promising first step towards demystifying these widely-used models.
 We show that with some knowledge of the input distribution, we can solve the extremely hard problem of training nonlinear IO-RNNs. We assume knowledge of score function forms, which correspond to normalized derivatives of the input probability density function (p.d.f). For instance, if the input is standard Gaussian, score functions are given by the Hermite polynomials.
There are many unsupervised approaches for estimating the score function efficiently, see Appendix~\ref{sec:sf-estimation}. To estimate the score function, one does not need to estimate the density, and this distinction is especially crucial for models where the normalizing constant or the {\em partition function} is intractable to compute. Guarantees have been derived for estimating  score functions of many flexible model classes such as infinite dimensional exponential families~\citep{sriperumbudur2013density}.  In addition, in many settings, we have control over designing the input distribution and our method is directly applicable.

We assume a Markovian model for the input sequence $\lbrace x_1, \dotsc, x_n \rbrace$ on a continuous state space. For a Markovian model, the score function only depends on the Markov kernel, and has a compact representation, as seen in Section~\ref{sec:sf-markov}. The method readily leads to higher order Markov chains. In the main paper we discuss the first order Markov chain for notation simplicity and discuss the extension in Appendix~\ref{appendix:remark}.

\textbf{Tensor decomposition: }We form   cross-moments between the output label and score functions  of the input. For a vector input, the first order score function is a vector, second order is a matrix, and higher orders corresponds to tensors. Hence, the empirical moments are tensors, and we  perform a CP tensor decomposition to obtain the rank-$1$ components. Efficient algorithms for tensor decomposition have been proposed before, based on simple iterative updates such as tensor power method~\citep{JMLR:v15:anandkumar14b}.  After some simple manipulations on the components, we provide estimates of the network  parameters of RNN models. The overall algorithm involves simple linear and multilinear steps,   is embarrassingly parallel, and is practical to implement~\citep{wang2015fast}. 
 %See~\citep{wang2015fast} for more discussion on algorithmic issues.

%It is competitive with standard backpropagation algorithm and requires only an $O(d_x)$ factor

\textbf{Recovery  guarantees: }We guarantee consistent recovery under (a low order) polynomial and computational complexity. We consider   the realizable setting when   samples are generated by a IO-RNN or a BRNN under the following transparent conditions:
(i) one hidden layer of neurons with a polynomial activation function, (ii) Markovian input sequence, (iii) full  rank weight matrices on input, hidden and output layers, and (iv) spectral norm bounds on the weight matrices to ensure bounded state evolution.

Currently,  the question of approximation bounds by a RNN with a fixed
number of neurons is not satisfactorily resolved~\citep{hammer2000approximation}  and it is valid to first consider the realizeable setting for this complex problem. The polynomial activations are a departure from the usual sigmoidal units, but they can also capture non-linear signal evolution, and have been employed in different applications, e.g.,~\citep{chen2014fast},~\citep{xie2015recurrent}. The Markovian assumption on the input limits the extent of dependence and allows us to derive concentration bounds for our empirical moments. %\tcr{Our results can be extended to  general order Markov models and latent space models such as HMMs.}
The full rank conditions on the weight matrices imply non-degeneracy in the neural representation: the weights for any two neurons cannot linearly combine to generate the weight of another neuron. Such conditions have previously been derived for spectral learning of HMMs and other latent variable models~\citep{JMLR:v15:anandkumar14b}. Moreover, it can be easily relaxed by considering higher order moment tensors, and is relevant when we want to have more neurons than the input dimension in our network. The rank assumption on the output weight matrix implies a vector output of sufficient dimensions, i.e., sufficient number of output classes. This can be relaxed to a scalar output, the details are given in Appendix~\ref{Appendix:scalar}.

The spectral norm condition on the weight matrices arises in the analysis of concentration bounds for the empirical moments. Since we assume polynomial state evolution, it is important to ensure bounded values of the hidden states, and this entails a bound on the spectral norm of the weight matrices. We employ  concentration bounds for functions of Markovian input from~\citep{kontorovich2008concentration,kontorovich2014uniform} and combine it with matrix Azuma's inequality~\citep{tropp2012user} to obtain concentration of empirical moment tensors. This implies learning RNNs with polynomial sample complexity.

\textbf{Related work: } The following works are directly relevant to this paper.
\textbf{(a) Spectral approaches for sequence learning: }Previous guaranteed approaches for sequence learning mostly focus on the class of hidden Markov models (HMM).~\citet{JMLR:v15:anandkumar14b} provide a tensor decomposition method for learning the parameters under non-degeneracy conditions, similar to ours. This framework is extended to more general HMMs in~\citep{huang2014minimal}.  While in a HMM, the relationship between the hidden and observed variables can be modeled as a linear one,
%FORNOW
%\footnote{Although the linear relationship is not obvious in a discrete model, this is clear under one hot encoding of both hidden variable $h$ and observed variable $y$, here $\Ebb[y|h]=Ah$, where $A$ is the conditional probability table, and $h$ selects a column of the table depending on the hidden state.}
  in a RNN it is non-linear. However, in a IO-RNN, we have both inputs and outputs, and that is helpful in handling the non-linearities.
\textbf{(b) Input-output sequence models: }A rich set of models based on RNNs have been employed in practice in a wide range of applications.~\citet{lipton2015critical} provides a nice overview of these various models.~\citet{BalduzziStrong} recently apply physics based principles to design RNNs for stabilizing gradients and getting better training error.  However, a rigorous analysis of these techniques is lacking.

%\paragraph{Score function formulation: } 

\section{Preliminaries} \label{notation2}

Let $[n] := \{1,2,\dotsc,n\}$, and
$\inner{u,v}$ denote the inner product of vectors $u$ and $v$. For sequence of $n$ vectors $z_1, \dotsc, z_n$, we use the notation $z{[n]}$ to denote the whole sequence.
For vector $v$, $v^{*m}$ refers to elementwise $m^{\tha}$ power of $v$.
For matrix $C \in \R^{d \times k}$, the $j$-th column is referred by $C_j$ or $c_j$, $j \in [k]$, the $j^{\tha}$ row is referred by $C^{(j)}$ or $c^{(j)}$, $j \in [d]$.
 Throughout this paper, $\nabla_x^{(m)}$ denotes the $m^{\tha}$ order derivative operator w.r.t.\ variable $x$. %and notation $\otimes$ represents tensor (outer) product. 	

\textbf{Tensor:}
A real \emph{$m^{\tha}$ order tensor} $T \in \bigotimes^m \R^d$ is a member of the outer product of Euclidean spaces $\R^{d}$.
The different dimensions of the tensor are referred to as {\em modes}.
% For instance, for a matrix, the first mode refers to columns and the second mode refers to rows.

\textbf{Tensor Reshaping: }
$T_2=\text{Reshape}(T_1,v_1, \dotsc, v_l)$ means that $T_2$ is a tensor of order $l$ that is made by  reshaping tensor $T_1$ such that the first mode of $T_2$ includes modes of $T_1$ that are shown in $v_1$,
the second mode of $T_2$ includes modes of $T_1$ that are shown in $v_2$ and so on. For example if $T_1$ is a tensor of order $5$,  $T_2=\text{Reshape}(T_1,[1~2],3 ,[4~ 5] )$ is a third order tensor, where its first mode is made by concatenation of modes $1, 2$ of $T_1$ and so on.

\textbf{Tensor rank: }A $3$rd order tensor $T \in \Rbb^{d \times d \times d}$ is said to be rank-$1$ if it can be written in the form
%\begin{equation*} \label{eqn:rank-1 tensor}
$T= w \cdot a \otimes b\otimes c \Leftrightarrow T(i,j,l) = w \cdot a(i) \cdot b(j) \cdot c(l),$
%\end{equation*}
where $\otimes$  represents the {\em outer product}, and $a, b , c \in \Rbb^d$ are unit vectors.
A tensor $T  \in \Rbb^{d \times d \times d}$ is said to have a CP (Candecomp/Parafac) {\em rank} $k$ if it can be (minimally) written as the sum of $k$ rank-$1$ tensors

$T = \sum_{i\in [k]} w_i a_i \otimes b_i \otimes c_i, \quad w_i \in \Rbb, \ a_i,b_i,c_i \in \Rbb^d.$
Note that $v^{\otimes p}= v \otimes v \otimes v \cdots \otimes v$, where $v$ is repeated $p$ times.

\begin{definition}[Row-wise Kronecker product] \label{KR}
For matrices $A, B \in \R^{d \times k}$, the {\em Row-wise Kronecker product} $\in \R^{d \times k^2}$ is defined as follows.

Let $a^{(i)}, b^{(i)}$ be rows of $A, B$ respectively. Rows of $A \odot B$ are of the form $a^{(i)} \otimes b^{(i)}$. Note that our definition is different from usual definition of Khatri-Rao product which is a column-wise Kronecker product. 
\end{definition}

\textbf{Derivative:}
For function $g(x): \R^d \rightarrow \R$ with vector input $x \in \R^d$, the $m$-th order derivative w.r.t.\ variable $x$ is denoted by $\nabla_x^{(m)} g(x) \in \bigotimes^{m} \R^d$ (which is a $m$-th order tensor) such that
\begin{equation*} \label{eqn:derivativedef}
\left[ \nabla_x^{(m)} g(x) \right]_{i_1,\dotsc,i_m}\! := \!\frac{\partial g(x)}{\partial x_{i_1} \partial x_{i_2} \dotsb \partial x_{i_m}}, ~ i_1,\dotsc,i_m \!\in\! [d].
\end{equation*}

\textbf{Tensor as multilinear form:} We view a tensor $T \in \Rbb^{d \times d \times d}$ as a multilinear form. %For simplicity let us consider order-3 tensors.
Consider matrices $M_l \in \R^{d\times d_l}, l \in \{1,2,3\}$. Then tensor $T(M_1,M_2,M_3) \in \R^{d_1}\otimes \R^{d_2}\otimes \R^{d_3}$ is defined as
\begin{align*} %\label{eqn:multilinear form def}
&T(M_1,M_2,M_3)_{i_1,i_2,i_3} := \\
&~\sum_{j_1, j_2,j_3\in[d]} T_{j_1,j_2,j_3} \cdot M_1(j_1, i_1) \cdot M_2(j_2, i_2) \cdot M_3(j_3, i_3).
\end{align*}
%In particular, if $u$, $v$ and $w$ are vectors and $T$ is a $3$rd order tensor, then $T(u,v,w)$ is a scalar, $T(I,v,w)$ is a vector, and $T(I, I, w)$ is a matrix.
In particular, for vectors $u,v,w \in \R^d$, we have\,\footnote{Compare with the matrix case where for $M \in \R^{d \times d}$, we have $ M(I,u) = Mu := \sum_{j \in [d]} u_j M(:,j) \in \R^d$.}
\begin{equation*} %\label{eqn:rank-1 update}
 T(I,v,w) = \sum_{j,l \in [d]} v_j w_l T(:,j,l) \ \in \R^d,
\end{equation*}
which is a multilinear combination of the tensor mode-$1$ fibers.
Similarly $T(u,v,w) \in \R$ is a multilinear combination of the tensor entries,  and $T(I, I, w) \in \R^{d \times d}$ is a linear combination of the tensor slices.

\subsection{Problem Formulation}

We consider a two-layer input-output RNN, that includes both regression and classification settings:
\begin{align*}
\Ebb[ y_t| h_t] = A_2^\top h_t, \quad
h_t =\poly_l (A_1 x_{t}+ U h_{t-1}),
\end{align*}
where $\poly_l(\cdot)$ denotes an element-wise polynomial of order $l$, The input sequence $x$ consists of the vectors $x_t \in \mathbb{R}^{d_x}. h_t \in \mathbb{R}^{d_h}, y_t \in \mathbb{R}^{d_y}$ and hence $A_1 \in \mathbb{R}^{d_h \times d_x}, U \in \mathbb{R}^{d_h \times d_h}$ and $A_2 \in \mathbb{R}^{d_h \times d_y}$. We can learn the parameters of the model using our method. Our algorithm is called GLOREE (Guaranteed Learning Of Recurrent nEural nEtworks) and is shown in Algorithm~\ref{algo:mainRNN}. 

Throughout the paper, we assume that the p.d.f. of the input sequence vanishes in the boundary (i.e., when any coordinate of the input goes to infinity). This is also the assumption in~\citep{janzamin2014matrix}.
We consider the case where input is a geometrically ergodic Markov chain. Then in order to have mixing and assure ergodicity for the output, we need to impose additional constraints on the model.

\subsection{Review of Score functions}
As mentioned in the introduction, our method  builds on the method introduced by~\citet{janzamin2014matrix} called FEAST (Feature ExtrAction using Score function Tensors). The goal of FEAST is to extract discriminative directions using   the cross-moment between the label and score  function of input. Score function is the normalized (higher order) derivative of p.d.f. of the input.

Let $p(x)$ denote the joint probability density function of random vector $x \in \R^d$. \citet{janzamin2014matrix} denote $\Pc_m(x)$ as the $m^{\tha}$ order score function,   given by
\begin{align} \label{eqn:highorder}
\Pc_m(x)=(-1)^m \frac{\nabla_x^{(m)} p(x)}{p(x)},
\end{align}
where $\nabla_x^{(m)}$ denotes the $m^{\tha}$ order derivative operator w.r.t.\ variable $x$. It   can also be derived using the recursive form
\begin{align} \label{eqn:scorerecursive}
&\Pc_1(x) = -\nabla_x \log p(x), \\ \nonumber
&\Pc_m(x) = -\Pc_{m-1}(x) \otimes \nabla_x \log p(x)  -\nabla_x \Pc_{m-1}(x).
\end{align}

The importance of score function is that it provides a derivative operator. ~\citet{janzamin2014matrix} proved that the cross-moment between the label and the score function of the input yields the information regarding derivative of the label w.r.t.\ the input.

\begin{theorem}[Yielding differential operators~\citep{janzamin2014matrix}] \label{thm:steins_higher}
Let $x \in \mathbb{R}^{d_x}$ be a random vector with joint density function $p(x)$. Suppose the $m^{\tha}$ order score function $\Pc_m(x)$ defined in~\eqref{eqn:highorder} exists.
Consider any order $m$ continuously differentiable tensor function $G(x):\mathbb{R}^{d_x} \rightarrow \bigotimes^r \mathbb{R}^{d_y}$. Then, under some mild regularity conditions\footnote{Consider any continuously differentiable tensor function $G(x):\mathbb{R}^{d_x} \rightarrow \bigotimes^r \mathbb{R}^{d_y}$ satisfying the regularity condition such that all the entries of $\nabla_x^{(i)} G(x) \otimes \Pc_{m-i-1}(x) \otimes p(x)$, $i \in \{0,1,\dotsc,m-1\}$, go to zero on the boundaries of support of $p(x)$.}, we have
\[
\Ebb \left[ G(x) \otimes \Pc_m(x) \right] = \\\Ebb \left[ \nabla^{(m)}_x G(x) \right].
\]
\end{theorem}

\section{Extension of score function to input sequences}

\subsection{Score function form for RNN}

We now extend the notion of score function to handle sequence data with non i.i.d. samples. We denote  the score function at each time step $t$ in the sequence as  $\Pc_m(z[n], t )$, where $z[n]:=z_1, z_2, \dotsc, z_n$, and it is defined below.
%The score function for the entire sequence as $\Pc_m(z[n], [n] ):= [ \Pc_m(z[n], 1)^\top, \Pc_m(z[n],2)^\top,\ldots]^\top$ is the concatenation of the scores at each time step.
 Theorem~\ref{thm:steins_higher} can be readily modified to
\begin{lemma}[Score function form for input sequence] \label{thm:steins_expand}
For vector sequence $z{[n]}=\lbrace z_1, z_2, \dotsc , z_n \rbrace$, let $p(z_1,z_2, \cdots, z_n)$ and $\Pc_m(z[n], [n])$ respectively denote the joint density function and the corresponding $m^{\tha}$ order score function. Then, under some mild regularity conditions,
%\footnote{any continuously differentiable tensor function $G(\cdot)$ satisfying the regularity condition such that all the entries of  $\nabla_x^{(i)} G(x) \otimes S_{m−i−1}(x) \otimes p(x), i \in \lbrace 0, 1, . . . , m -1\rbrace,$ go to zero on the boundaries of support of $p(x)$. }
 for all continuously differentiable functions $G(z_1,z_2, \dotsc, z_n)$, we have
\begin{align*}
\Ebb \left[ G(z_1,z_2, \dotsc, z_n) \otimes \Pc_m(z[n], t) \right] = \Ebb \left[ \nabla^{(m)}_{z_t} G(z_1,z_2, \dotsc, z_n) \right],
\end{align*}
where $\nabla_{z_t}^{(m)}$ denotes the $m^{\tha}$ order derivative operator w.r.t.\ $z_t$,%\footnote{Note the difference from i.i.d. case, for details see Appendix~\ref{appendix:RNN-Q}.},
\begin{align}
\label{eqn:highorderexpand}
 \Pc_m(z[n], t) = (-1)^m \frac{\nabla_{z_t}^{(m)} p(z_1,z_2, \dotsc, z_n)}{p(z_1,z_2, \dotsc, z_n)}.
\end{align}
\end{lemma}

%\paragraph{Remark: }In the current paper, we partition the input sequence $x_i, i \in [t]$ into $x_t, x_{\setminus t}$. Therefore, when we fare looking for $\Ebb \left[ \nabla^{(m)}_{z_1} G(z_1,z_2) \right]$, we form the complete tensor $\Ebb \left[ G(z_1,z_2) \otimes \Pc_m(z_1,z_2) \right]$ and then look at the section corresponding to $z_1$. This is the convention we will use throughout the paper.

\subsection{Score function form for Markov chains} \label{sec:sf-markov}
We assume a Markovian model for the input sequence, and derive compact score function forms for~\eqref{eqn:highorderexpand}.   Note that this form can be readily expanded to higher-order Markov chains.  

\begin{lemma}[Score function for first-order Markov Chains]
Let the input sequence $\lbrace x_i \rbrace_{i \in [n]}$ be a first-order Markov chain. The score function in \eqref{eqn:highorderexpand} simplifies as 
\begin{align}
\label{eqn:sf-mc}
\Pc_m(x[n], i)=(-1)^m \frac{\nabla_{x_i}^{(m)} \left[p(x_{i+1}|x_i)p(x_i|x_{i-1})\right]}{p(x_{i+1}|x_i)p(x_i|x_{i-1})}.
\end{align}
\end{lemma}
The proof follows the definition of first-order Markov chain and Equation~\eqref{eqn:highorderexpand}.

%\textbf{Score function form for AR processes}
%The Gaussian AR-$1$ process is a special case of the Markov chain: we have $ x_t=\phi x_{t-1}+\epsilon_t, \quad \epsilon_t \sim \mathcal{N}(0,1), ~~|\phi| < 1,$
%where index $t$ represents the time and $|\phi|<1$. For simplicity, we have considered scalar $x$, the form can be easily extended to vector $x$.
%
%The conditional distribution is given by 
%$x_t | x_1, \cdots, x_{t-1} \sim \mathcal{N}(\phi x_{t-1},1),  t=2, \cdots, n, $
%and~\citep{rue2005gaussian}
%\begin{align*}
%&p(x[n]) = \frac{1}{(2\phi)^{n/2}} |J|^{1/2} \exp\left(-\frac{1}{2}x[n]^\top J x[n]\right),\\
%&J_{ij}=\left\lbrace 
%\begin{array}{ll}
%-\phi & i=j+1,i=j-1\\
%1+\phi^2 & i=j\neq 1\\
%1 & i=j=1\\
%0 & o.w. \end{array}\right.
%\end{align*}
%
%Expanding Equation~\eqref{eqn:scorerecursive} for any Gaussian AR process, we have that
%\begin{align} \label{eqn:AR1}
%&\Pc_1(x[n], [n]) = x[n]^\top J, \\ \nonumber
%&\Pc_m(x[n], [n]) =  \\ \nonumber
%&\quad \Pc_{m-1}(x[n], [n]) \otimes (x[n]^\top J) - \nabla_{x[n]} \Pc_{m-1}(x[n], [n]).%\nonumber
%\end{align}
%i.e.,
%\begin{align*}
%&\Pc_1(x[n], i) = \!\left\lbrace
%\begin{array}{ll}
%x_1-\phi x_2 & i=1\\
%-\phi x_{i-1}+(1+\phi^2)x_{i}-\phi x_{i+1} & i\neq1,n\\
%-\phi x_{n-1}+x_n & i=n\\
%\end{array}
%\right. 
%\end{align*}
%
%

\section{Algorithm and Guarantees}

In this paper, we have functions which map input sequence $x_1,\ldots x_n$ to an output sequence $y_1,\ldots, y_n$. By assuming a structured form of function mapping in terms of IO-RNN, we can hope to recover the function parameters efficiently.   We exploit the score function forms derived above to    compute partial derivatives  of the output sequence.%of the form $\Ebb[\nabla^m_{x_{\tau_1}\ldots x_{\tau_m}} y_t]$.
  We first start with some simple intuitions.

\subsection{Preliminary insights}

\textbf{Generalized linear model: }Before considering the RNN, consider a simple generalized linear model with i.i.d. samples:    $\Ebb[y | x ]=A_2^\top \sigma(A_1x )$, where $A_1$ is the weight matrix and $\sigma(\cdot)$ is element-wise activation.   Here, the partial derivative   of $\Ebb[y | x ]$ w.r.t.\ $x $  has a linear relationship with the weight matrices $A_1$ and $A_2$, i.e.,
\begin{subequations}
\begin{align}
&\Ebb[ \nabla_{x }\Ebb[y| x]]= \Ebb[A_2^\top\nabla_{x} \sigma(A_1x )]=A_2^\top\Ebb[\sigma'(A_1x )]A_1.
\label{eqn:glm}\\
&\Ebb[ y \otimes \Pc_2(x)] = \Ebb[\nabla_{x}^2 \Ebb[  y | x ]]= \sum_{i \in d_h} \mu_i A_2^{(i)} \otimes A_1^{(i)} \otimes A_1^{(i)}.\label{eqn:glm2}
\end{align}
\end{subequations}
The first partial derivative  is obtained by forming the cross moment $\Ebb[ y \otimes \Sc_1(x)]$, as given by Theorem~\ref{thm:steins_higher}.
The form in \eqref{eqn:glm} yields $A_1$ and $A_2$ up to a linear transformation. But, by computing second order derivatives, we can recover $A_1$ and $A_2$, up to scaling of their rows.   The second order derivative has the form in~\eqref{eqn:glm2}.
%\beq \label{eqn:glm2}\Ebb[ y \otimes \Pc_2(x)] = \nabla_{x}^2 \Ebb[  y | x ]= \sum_{i \in d_h} \mu_i A_2^{(i)} \otimes A_1^{(i)} \otimes A_1^{(i)}.\eeq
The  tensor decomposition in~\citep{JMLR:v15:anandkumar14b} uniquely recovers  $A_1, A_2$ up to scaling of rows, under full row rank assumptions.

\textbf{Recovering input-output weight matrices in IO-RNN: }The above intuition for GLM readily carries over to IO-RNN. Recall that the IO-RNN has the form \begin{align*}
\Ebb[ y_t| h_t] = A_2^\top h_t, \quad
h_t = \poly_l(A_1 x_{t}+ U h_{t-1}),
\end{align*} where $\poly_l$ denotes any polynomial function of degree at most $l$.

Suppose we have access to partial derivatives $\Ebb[\nabla_{x_t}\Ebb[ y_t| h_t]]$ and $\Ebb[\nabla^2_{x_t}\Ebb[ y_t| h_t]]$, then they have   the same forms as \eqref{eqn:glm} and  \eqref{eqn:glm2}.\footnote{ Note that $h_t$ is a (polynomial) function of $x_t$ so $\Ebb[y_t|h_t]=poly(x_t, h_{t-1})$. Also  the expectation is w.r.t. all variables $x_1, \dotsc, x_t$, and thus, the dependence to $h_{t-1}$ is also averaged out since it is a function of $x_1, \dotsc, x_{t-1}$.} This is because $h_t$ does not depend on $x_{t-1}$ given $x_t, h_{t-1}$. Thus,  the weight matrices $A_1$ and $A_2$ can be easily recovered by forming $\Ebb[y_t \otimes \Sc_2(x[n],t)]$, as given by \eqref{eqn:highorderexpand}, and it has a compact form for Markovian input in \eqref{eqn:sf-mc}. Note that this intuition holds for any non-linear element-wise activation function, and we do not require it to be a polynomial at this stage.

\textbf{Recovering hidden state transition matrix in IO-RNN: }Recovering the transition matrix $U$ is much more challenging as we do not have access to hidden state sequence $h[n]$. Thus, we cannot readily form partial derivatives of the form $\nabla^m_{h_{t-1}}\Ebb[ y_t | h_{t}]$. Also, the non-linearities get recursively propagated along the chain. Here, we provide an algorithm that works for any polynomial activation
function  of fixed degree $l$.

The main idea is that we attempt to recover $U$ by considering partial derivatives $\nabla^m_{x_{t-1}}\Ebb[ y_t|h_{t}]$, i.e., how the previous input $x_{t-1}$ affects current output $y_t$.  At first glance, this appears complicated and indeed, the various terms are highly coupled and we do not have a nice CP tensor decomposition form. However, we can prove that when the derivative order $m$ is sufficiently large, a nice CP tensor form emerges out, and this $m$ depends on the degree $l$ of the polynomial activation.

For simplicity, we provide intuitions  for   the quadratic activation function $(l=2)$. Now, $y_t$ is a degree-$4$ polynomial function  of $x_{t-1}$, since the activation function is applied twice. By considering fourth order derivative $\nabla^4_{x_{t-1}}\Ebb[ y_t| h_{t}]$, many coupled terms vanish since they correspond to polynomial functions of degree less than $4$. The surviving term has a nice CP tensor form, and we can recover $U$ efficiently from it. Note that this fourth order partial derivative can be computed efficiently using fourth order score function and forming the cross-moment $\Ebb[y_t \otimes \Pc_{4}(x[n],{t-1})]$.  The precise algorithm is given in Algorithm~\ref{algo:mainRNN}.
\subsection{Training IO-RNNs }

We now provide our algorithm for training IO-RNNs. In this paper we consider vector output and polynomial activation functions of any order $l \geq 2$.   For simplicity of notation, we first discuss the case for quadratic activation function. %and show the general case of polynomials of order $l \geq 2$ in Appendix~\ref{appendix:RNN-Alg}. 
Our algorithm is called GLOREE (Guaranteed Learning Of Recurrent nEural nEtworks) and is shown in~Algorithm~\ref{algo:mainRNN} for quadratic activation function. The general algorithm and analysis for $l \geq 2$ is shown in Appendix~\ref{appendix:RNN-Alg}.
For completeness, we handle the linear case in Appendix~\ref{sec:linear}.  We also cover the case where output $y$ is a scalar (e.g. a binary label)  in Appendix~\ref{Appendix:scalar}.

\begin{algorithm}[t]
\caption{GLOREE (Guaranteed Learning Of Recurrent nEural nEtworks) for vector input and quadratic function (General case shown in Algorithm~\ref{algo:mainRNN_poly} in Appendix~\ref{appendix:RNN-Alg}).}
\label{algo:mainRNN}
\begin{algorithmic}[1]
\renewcommand{\algorithmicrequire}{\textbf{input}}
\renewcommand{\algorithmicensure}{\textbf{output}}
\REQUIRE Labeled samples $\{(x_i,y_i): i \in [n]\}$ from IO-RNN model in Figure~\ref{fig:RNN}.

\STATE Compute $2$nd-order score function $\Pc_2(x[n],i)$ of the input sequence as in Equation~\eqref{eqn:sf-mc}.
\STATE Compute $\widehat{T}:= \widehat{\Ebb}\left[ y_i \otimes \Pc_2(x[n], i) \right]$. The empirical average is over a single sequence.
%$\widehat{D}:= \widehat{\Ebb}\left[ y_i \otimes \Pc_1(x[n], i) \right]$. and $\widehat{D}$ will be used for whitening (Procedure~\ref{algo:whitening} in the Appendix).

\STATE $\lbrace \hat{w}, \hat{R}_1, \hat{R}_2,\hat{R}_3  \rbrace =\text{tensor decomposition}(\widehat{T})$; see Appendix~\ref{sec:tensor}. %\label{line:TensorDecomposition}
\STATE $\hat{A}_2=\hat{R}_1,   \hat{A}_1= \left(  \hat{R}_2 + \hat{R}_3 \right)/2 $.

\STATE Compute $4$th-order score function $\Pc_{4}(x[n], i)$ of the input sequence  as in Equation~\eqref{eqn:sf-mc}.

\STATE Compute $\widehat{T}_2:=\hat{\Ebb} \left[y_t\otimes \text{Reshape}\left((\Sc_{4}(x[n],{t-1}), [1 ~~ 2], [3~~4] \right)\right]$.

\STATE $\lbrace \hat{w}, \hat{R_1}, \hat{R_2}, \hat{R_{3}}  \rbrace =\text{tensor decomposition}(\widehat{T}_2)$; see Appendix~\ref{sec:tensor}.
%\STATE $\tl{R}=\left(  \hat{R}_2 + \hat{R}_3 \right)/2$.
\STATE $\hat{U}=\tl{R}\left[\hat{A}_1 \odot \hat{A}_1  \right]^{\dagger}$, $\tl{R}=\left(  \hat{R}_2 + \hat{R}_3 \right)/2$. $\odot$ is row-wise Kronecker product as in Definition~\ref{KR}.

\RETURN $\hat{A}_1, \hat{A}_2, \hat{U}$.

\end{algorithmic}
\end{algorithm}
%\lipsum[3-6]
%\setlength{\textfloatsep}{0pt}

Now, we consider an RNN with quadratic activation function and vector output. Let
\begin{align}
\label{IORNN}
\Ebb[ y_t| h_t] = A_2^\top h_t, \quad
h_t = \poly_2(A_1 x_{t}+ U h_{t-1}),
\end{align}
where $x_t \in \mathbb{R}^{d_x}, h_t \in \mathbb{R}^{d_h}, y_t \in \mathbb{R}^{d_y}$ and  $A_1 \in \mathbb{R}^{d_h \times d_x}, U \in \mathbb{R}^{d_h \times d_h}$, $A_2 \in \mathbb{R}^{d_h \times d_y}$. We can learn the parameters of the model using GLOREE. Let $n$ be the window size for RNN. 

\begin{theorem}[Learning parameters of RNN for quadratic activation function] \label{lemma:RNN-Q} 
Let $\mathcal{R}$ be the model describing the IO-RNN as in~\eqref{IORNN}. Assuming that $A_1, A_2, U$ are full column rank, we can recover the parameters of $\mathcal{R}$ using  Algorithm~\ref{algo:mainRNN} (GLOREE).
\end{theorem}

\textbf{Proof Sketch: }We have the following properties for an IO-RNN:
\begin{align}\label{eqn:quad}
\Ebb \left[ y_t \otimes \Sc_2(x[n], t) \right] = 2 \sum_{i \in d_h} A_2^{(i)} \otimes A_1^{(i)} \otimes A_1^{(i)}.
\end{align}
Hence, we can recover $A_1, A_2$ via tensor decomposition, assuming that they are full row rank.
%\end{lemma}

In order to learn $U$ we form the tensor $\Ebb \left[ y_t \otimes \Sc_4(x_{t-1}) \right]$, and under quadratic activations, we have
%\begin{lemma}
\begin{align*}
\Ebb \left[ y_t\otimes \text{Reshape}(\Sc_4(x[n],{t-1}),[1~ 2], [3~ 4]) \right] =  
\sum_{i \in d_h} A_2^{(i)} \otimes [U(A_1 \odot A_1)]^{(i)} \otimes [U(A_1 \odot A_1)]^{(i)}.
\end{align*}
Hence, we can recover $U(A_1 \odot A_1)$ via tensor decomposition. Since $A_1$ is previously recovered using \eqref{eqn:quad}, and $(A_1\odot A_1)^\dagger$ exists due to full rank assumption, we can recover $U$. Thus, Algorithm~\ref{algo:mainRNN} (GLOREE) consistently recovers the parameters of IO-RNN under quadratic activations. For proof, see Appendix~\ref{appendix:RNN-Q}.

%\bprf The underlying idea behind the proof comes from Theorem~\ref{thm:steins_higher}. Hence
%%\begin{proof}
%%By Theorem~\ref{thm:steins_higher} we have that
%\begin{align*}
%\Ebb \left[ y_t \otimes \Sc_2(x[n], t) \right] &= \nabla^2_{x_t} \Ebb \left[  y_t| x_t  \right], \\
%\Ebb \left[ y_t \otimes \text{Reshape}(\Sc_4(x[n],{t-1}),[1 ~2], [3~ 4]) \right] &= \tcr{\text{Reshape}(\nabla^4_{x_{t-1}}\Ebb \left[  y_t, |x_{t-1}\right],[1~2], [3~ 4]).}
%\end{align*}
%\begin{align*}
%T= \nabla^4_{x_{t-1}}\Ebb \left[ y_t |x_{t-1}\right]&=2\sum_{j \in d_h} A_2^{(j)} \otimes \sum_{k \in d_h} U_{jk}A_1^{(k)} \otimes A_1^{(k)} \otimes  \sum_{m \in d_h} U_{jm}A_1^{(m)} \otimes A_1^{(m)},
%\end{align*}
%By reshaping the above tensor to \tcr{$\text{Reshape}( \nabla^4_{x_{t-1}} y_t, [1~2], [3~ 4])$ we have the form $\sum_{j \in d_h} A_2^{(j)} \otimes [U(A_1 \odot A_1)]^{(j)} \otimes [U(A_1 \odot A_1)]^{(j)}.$}
 %\eprf

\begin{algorithm}[t]
\caption{GLOREE-B (Guaranteed Learning Of Recurrent nEural nEtworks-Bidirection case) for quadratic activation function (general case is shown in Algorithm~\ref{algo:BRNN-general}).}
\label{algo:BRNN}
\begin{algorithmic}[1]
\renewcommand{\algorithmicrequire}{\textbf{input}}
\renewcommand{\algorithmicensure}{\textbf{output}}
\REQUIRE Labeled samples $\{(x_i,y_i): i \in [n]\}$ from~\eqref{eqn:BRNN}.
%, both activation functions in the forward and backward direction are quadratic.

\REQUIRE $2$nd-order score function $\Pc_2(x[n], [n])$ of the input $x$; see Equation~\eqref{eqn:highorder} for the definition.

\STATE Compute $\widehat{T}:= \hat{\Ebb} [y_i \otimes \Pc_2(x[n], i)]$. The empirical average is over a single sequence.
% $\widehat{D}:= \hat{\Ebb} [y_i \otimes \Pc_1(x[n], i)]$.  and $\widehat{D}$ will be used for whitening (Procedure~\ref{algo:whitening}).
\STATE $\lbrace (\hat{w}, \hat{R}_1, \hat{R}_2,\hat{R}_3  \rbrace=\text{tensor decomposition}(\widehat{T})$; see Appendix~\ref{sec:tensor}.

\STATE $\hat{A}_2 = \hat{R}_1 $.
\STATE  Compute $\tl{T}=\widehat{T}(((\hat{A}_2)^\top)^{-1}, I, I)$. For definition of multilinear form see Section~\ref{notation2}.
\STATE $\lbrace (\hat{w}, \hat{R}_1, \hat{R}_2,\hat{R}_3  \rbrace=\text{tensor decomposition}(\tl{T})$;
\STATE $   \hat{C}=(\hat{R}_2+\hat{R}_3)/2 $.
\STATE $\hat{C}=\left[\begin{array}{c}
\hat{A}_1 \\
\hat{B}_1
\end{array}\right] $.

\STATE Compute $4$th-order score function $\Pc_{4}(x[n],{t-1})$ of the input sequence  as in Equation~\eqref{eqn:sf-mc}. \label{line:beginQ}

\STATE Compute  $ \widehat{T} =\hat{\Ebb} \left[\!y_t\otimes \text{Reshape}\left((\Sc_{4}(x[n], {t-1}),[1 ~2], [3~4]\right) \right]$.
%\STATE \hscomment{change} Contract the moment-tensor to a $3$-rd order tensor as $\widehat{T}_2=T(I,  \theta,\theta)$ with some random vector $\theta$.
%\STATE Compute $\widehat{T}:= \frac{1}{n} \sum_{i \in [n]} y_i \cdot \Pc_2(x_i)$. %, Empirical estimate of 
\STATE $\lbrace \hat{w}, \hat{R_1}, \hat{R_2}, \hat{R_{3}}  \rbrace =\text{tensor decomposition}(\widehat{T})$; see Appendix~\ref{sec:tensor}.
%\STATE $\tl{R}=\left(  \hat{R}_2 + \hat{R}_3 \right)/2$.
\STATE $\hat{U}=\tl{R}\left[\hat{A}_1 \odot \hat{A}_1 \right]^{\dagger}$,  $\tl{R}=\left(  \hat{R}_2 + \hat{R}_3 \right)/2$. $\odot$ is row-wise Kronecker product as in Definition~\ref{KR}. \label{line:endQ}
%\STATE Compute $\widehat{T}:= \frac{1}{n} \sum_{i \in [n]} y_i \cdot \Pc_2(x_i)$. %, Empirical estimate of 
%\STATE Compute $\widehat{T}=\hat{\Ebb} \left[y_t\otimes \text{Reshape}(\Sc_{4}(x[n],{t+1}),1, [1 ~ 2],\dotsc, [3~4] \right]$.
%\STATE Contract the moment-tensor to a $3$-rd order tensor as $\widehat{T}_2=T(I, I, I, \theta, \theta)$ where $\theta \sim \mathcal{N}(0,I_{d})$ is a random standard Gaussian vector.
%%\STATE Compute $\widehat{T}:= \frac{1}{n} \sum_{i \in [n]} y_i \cdot \Pc_2(x_i)$. %, Empirical estimate of 
%\STATE $\lbrace \hat{w}, \hat{R_1}, \hat{R_2}, \hat{R_{3}}  \rbrace =\text{tensor decomposition}(\widehat{T}_2)$; see Algorithm~\ref{algo:main} in the Appendix.
%\STATE $\tl{R}=\left(  \hat{R}_2 + \hat{R}_3 \right)/2$.
%\STATE $\hat{V}=\tl{R}\left[\hat{B}_1 \odot \hat{B}_1 \right]^{-1}$, Khatri-Rao product $\odot$ is defined in Definition~\ref{KR}.
\STATE Repeat lines~\eqref{line:beginQ}-\eqref{line:endQ} with $\Pc_{4}(x[n],{t+1})$ instead of $\Pc_{4}(x[n],{t-1})$ to recover $\hat{V}$.

%\STATE Compute $u_1$ and $v_1$ as the top left and right singular vectors of  $T(I,I,\theta) \in \R^{d \times d}$.
%\STATE $\ha^{(0)} \leftarrow u_1$, $\hb^{(0)} \leftarrow v_1$.
%\STATE Initialize $\hc^{(0)}$ by update formula in \eqref{eqn:asymmetric power update}.
\RETURN $\hat{A}_1, \hat{A}_2, \hat{B}_1, \hat{U}, \hat{V}$.

\end{algorithmic}
\end{algorithm}
%\setlength{\textfloatsep}{0pt}
%\subsection{Bidirectional RNN with quadratic activation functions}
\subsection{Training Bidirectional RNNs}

Bidirectional Recurrent Neural network  was first proposed by~\citet{schuster1997bidirectional}. Here there are two groups of hidden neurons.The first group receives  recurrent connections from previous time steps while the other from the next time steps. The following equations describe a BRNN
\begin{align}
\label{eqn:BRNN}
\Ebb[y_t| h_t, z_t ] = A_2^\top \left[\begin{array}{c} h_t \\ z_t
\end{array} \right], \quad
h_t  = f(A_1 x_t + U h_{t-1}), \quad
z_t  = g(B_1 x_t + V z_{t+1}),
\end{align}
where $h_t$ and  $z_t$ denote the neurons that receive forward and backward connections, respectively.  Note that BRNNs cannot be used in online settings as they require knowledge of the future steps. However, in various natural language processing applications such as part of speech BRNNs are effective models since they consider both past and future words in a sentence.

We can learn the parameters of bidirectional RNN by modifying our earlier algorithm.
%Due to lack of space, the Algorithm is shown in the Appendix. 
For notation simplicity, Algorithm~\ref{algo:BRNN} shows the case for quadratic activation functions $f(\cdot)$ and $g(\cdot)$. The general polynomial function is considered in Algorithm~\ref{algo:BRNN-general} in Appendix~\ref{appendix:BRNNg}. 
%Due to lack of space, this is shown in Algorithm~\ref{algo:BRNN} in the Appendix~\ref{appendix:BRNNg1}. For simplicity, we limit ourselves to  quadratic activation functions for $f(\cdot)$ and $g(\cdot)$  here. The general polynomial function is shown in Algorithm~\ref{algo:BRNN-general} in Appendix~\ref{appendix:BRNNg}.

Let us provide some intuitions. If we only had forward or backward connections, we would directly apply our previous method in GLOREE. For backward connections, the only difference would be to use derivatives of $\Ebb[y_t| h_t, z_t]$ w.r.t.\ $x_{t+1}$ to learn the transition matrix $V$.
Now that we have both hidden neurons mixing to yield the output vector $y_t$,   the cross-moment tensor $T = \Ebb[y_i \otimes \Pc_2(x[n], i)]$ has a CP decomposition where the factor matrix for the first mode is $A_2$, i.e., the tensor has a specific form: $T= A_2^\top\left[\begin{array}{c} T_h \\ T_z \end{array} \right]$, where $T_h$ corresponds to the tensor incorporating  columns of $A_1$  and $T_{z}$ incorporates  columns of $B_1$. Hence, under full rank assumption, as before, we can recover $A_2$.  Next, we can invert $A_2$ to recover $T_h$ and $T_z$. We decompose them to recover $A_1$ and $B_1$.
%\tcr{
The steps for  recovering $U$ and $V$ remains the same as before in GLOREE. The only difference is to use derivatives of $\Ebb[y_t| h_t, z_t]$ w.r.t.\ $x_{t+1}$ to learn $V$. 
%Note that in order to recover $U$ and $V$, we can either decompose a tensor of order $l+1$ or contract the tensor to a third-order tensor, and then decompose the contracted tensor. For simplicity, we show the latter in our algorithm.}

\begin{theorem}[Training BRNN] \label{lemma:BRNN}
Let $\mathcal{B}$ be the BRNN model in~\eqref{eqn:BRNN}. Assuming that $A_1, A_2, B_1, U, V$ are full column rank, we can recover the parameters of $\mathcal{B}$ using  Algorithm~\ref{algo:BRNN}.
 \normalfont For proof, see Appendix~\ref{appendix:proofsBRNN}.
\end{theorem}

%For proof, see Appendix~\ref{appendix:proofsBRNN}.

\subsection{Analysis of GLOREE}

\subsubsection*{Sample Complexity: } \label{main:sample}
In order to analyze the sample complexity, we first need to prove concentration bound for the cross-moment tensor, then we use analysis of the tensor decomposition to show the that sample complexity is a low order polynomial of corresponding parameters.  
%Deriving concentration bound for the cross-moment tensor is a bit more involved since we have a non-i.i.d. sequence. 
%Moreover, since we assume a  polynomial activation function, the hidden state $h_t$ can grow in an unbounded manner. To avoid this, we impose additional assumptions on spectral norm of the model matrices and without loss of generality assume that $\ell_2$ norm of each entry in the input sequence is bounded by 1. 

\textbf{Assumptions: }\begin{enumerate} \item \textbf{Bounded hidden variables :} since we assume a  polynomial activation function, the hidden state $h_t$ can grow in an unbounded manner. To avoid this we need the following: \begin{enumerate} \item Without loss of generality, we assume that the input sequence is bounded by $1$ with high probability, $\| x_i \| <1~~ \forall i \in [n]$ \item Assume that $\| A_1 \| + \| U \| \leq 1$. \item $\Vert A_2 \Vert$ is bounded.
\end{enumerate}
\item \textbf{Concentration of moment tensor: } Deriving concentration bound for the cross-moment tensor is a bit more involved since we have a non-i.i.d. sequence. 
\begin{enumerate}
\item The input sequence is a geometrically ergodic Markov chain.
\item If the activation function is a polynomial of order $l$, we need $\| U\| \leq 1/l$.
\item $\Vert\Pc_2(x[n], t)\Vert$ is a bounded value.
\item The input sequence is a first order Markov chain.
\item $ \| \nabla_{x_{i}}\Pc_2(x[n], t) \|, i \in \lbrace t-1 , t, t+1 \rbrace$ is bounded by some value $\gamma$.\footnote{ We need milder assumptions than 2(d)-2(e).  For details see Appendix~\ref{appendix:remark}.} 
%$\left\lbrace\|\Pc_3(x[n], i) - \Pc_2(x[n], i) \otimes \Pc_1(x[n], i)\|\right\rbrace$ is bounded.
%\hscomment{need to change the assumption form for derivative of score function}
\end{enumerate} 

\item \textbf{Uniqueness in tensor decomposition: }  \begin{enumerate}
\item weight matrices $A_1, U, A_2$ are full column rank, i.e., neurons are not redundant.
\end{enumerate} \end{enumerate} 

%\textbf{Remark: } Note that wlog. we used $1$ as an upper bound in 1(a)-1(b). We can use any other finite number. In fact we need milder assumptions than 2(d)-2(e). For detailed analysis, see Appendix~\ref{sec:samplecomp}.

Let $G$ be the geometric ergodicity and $\theta$ is the concentration coefficient of the input Markov chain (see Appendix~\ref{sec:samplecomp} for definition). 
We have that:
\begin{theorem}[Sample Complexity for GLOREE] Assume the conditions above are met.
Suppose the sample
complexity $n$ is
\begin{align*}
\tilde{O}(d_x, d_y, d_h, G, \epsilon^{-2},\frac{1}{1-\theta}, \sigma_{\min}^{-1}(A_1), \sigma_{\min}^{-1}(A_2), \sigma_{\min}^{-1}(U)),
\end{align*}
then for each weight matrix column $A_1^{(i)} , U^{(i)} , A_2^{(i)}, i \in [d_h] $, we have that
\begin{align*}
\Vert \widehat{A}_1^{(i)} - A_1^{(i)} \Vert \leq \epsilon,\quad i \in [d_h], \\
\Vert \widehat{U}^{(i)} - U^{(i)} \Vert \leq \epsilon,\quad i \in [d_h] , \\
\Vert \widehat{A}_2^{(i)} - A_2^{(i)} \Vert \leq \epsilon,\quad i \in [d_h] .
\end{align*}
\end{theorem}
%\begin{theorem}[Sample Complexity for GLOREE] Assume the conditions above are met. In order to recover the weight matrix columns $A_1^{(i)} , U^{(i)} , A_2^{(i)}, i \in [d_h] $, with error $\epsilon \in (0,1)$,
%\begin{align*}
%\Vert \widehat{A}_1^{(i)} - A_1^{(i)} \Vert \leq \epsilon,\quad i \in [d_h], \\
%\Vert \widehat{U}^{(i)} - U^{(i)} \Vert \leq \epsilon,\quad i \in [d_h] , \\
%\Vert \widehat{A}_2^{(i)} - A_2^{(i)} \Vert \leq \epsilon,\quad i \in [d_h],
%\end{align*}
%we need $n$ samples no more than 
%\begin{align*}
%\tilde{O}(d_x, d_y, d_h, G, \epsilon^{-2},\frac{1}{1-\theta}, \sigma_{\min}^{-1}(A_1), \sigma_{\min}^{-1}(A_2), \sigma_{\min}^{-1}(U)).
%\end{align*}
%\end{theorem}

\paragraph{Proof Sketch: } The proof has two main parts. First, we need to prove a concentration bound for the moment tensor. Second, we can readily use the analysis of tensor decomposition from earlier works such as ~\citep{JMLR:v15:anandkumar14b,JanzaminEtal:NN2015} to compute the sample complexity for this moment tensor. Since the first part is the contribution of this paper, here we focus on that. 

In order to prove the concentration bound for the moment tensor, note that our input sequence $x[n]$ is a geometrically ergodic Markov chain. We can think of  the empirical moment $\hat{\Ebb}[y_t\otimes \Sc_m(x[n], t)]$ as functions  over the samples $x[n]$ of the Markov chain. Note that this assumes $h[n]$ and $y[n]$ as deterministic functions of $x[n]$, and our analysis can be extended when there is additional randomness.~\citet{kontorovich2014uniform} provide the result for scalar functions and this is an extension of that result to matrix-valued functions. We use Assumptions 1(a)-(c) to ensure a bounded hidden variable. Next, by leveraging Assumptions 2(a)-2(e) we prove that the cross-moment tensor satisfies Lipschitz property, which paves the way for proving the concentration bound.
For details, see Appendix~\ref{sec:samplecomp}.

\subsubsection*{Computational Complexity: }
The computational complexity of our method is related to the complexity of the tensor decomposition methods. See~\citep{JMLR:v15:anandkumar14b,JanzaminEtal:NN2015} for a detailed discussion. It is popular to perform the tensor decomposition in a stochastic manner   by splitting the data into mini-batches and reducing computational complexity. Starting with the first mini-batch, we perform a small number of tensor power iterations, and then use the result as initialization for the next mini-batch, and so on. We assume that score function is given to us in an efficient form. Note that if we can write the cross-moment tensor in terms of summation of rank-1 components, we do not need to form the whole tensor explicitly. As an example, if input follows Gaussian distribution, the score function has a simple polynomial form, and the computational complexity of tensor decomposition is $O(nd_hd_xR)$, where $n$ is the number of samples and $R$ is the number of initializations for the tensor decomposition. Similar argument follows when the input is mixture of Gaussian distributions.

\section{Conclusion}

This work is a first step towards answering challenging questions in sequence modeling. We propose the first method that can recover parameters of IO-RNN as well as BRNN with guarantees. Many of the assumptions can be relaxed, e.g., here we assumed  IO-RNNs with aligned inputs and outputs. We can relax this assumption to obtain more general RNNs. %As mentioned before, the framework directly extends to higher order Markov chains. 

This paper opens up a new horizon for future research,
% We have assumed polynomial activation functions at the neurons and our computational and sample complexity degrade exponentially in the degree of the polynomial. It is an open question to develop strategies to avoid this exponential blowup, and to extend it to  the usual sigmoidal units in RNNs. 
%Architectures such as long short-term memory (LSTM) have much more complicated non-linear dependencies, and it is unclear on how to analyze them effectively.
such as extending this framework to HMMs and general settings and  analysis under non-stationary inputs% is subject of future research.
% and involves more complicated score function forms. 
% Analysis under non-stationary inputs is another challenging open problem. 
We have assumed the realizable setting where samples are generated from a RNN.  The question of approximation bounds by a RNN with a fixed number of neurons is an interesting problem. 

\subsection*{Acknowledgment}
The authors thank Majid Janzamin for discussions on sample complexity and constructive comments on the draft. We thank Ashish Sabharwal for editorial comments on the draft.
This work was done during the time H. Sedghi was a visiting researcher at University of California, Irvine and was supported by NSF Career award FG15890. 
A. Anandkumar is supported in part by Microsoft Faculty Fellowship,
NSF Career award CCF-1254106,  ONR award N00014-14-1-0665, ARO YIP
award W911NF-13-1-0084, and AFOSR YIP award  FA9550-15-1-0221.

%\subsubsection*{References}

\appendix

\section{Notation} \label{sec:notation}
For completeness, all the notation required in the paper and Appendices are gathered here as well.

Let $[n] := \{1,2,\dotsc,n\}$, and $\|u\|$ denote the $\ell_2$ or Euclidean norm of vector $u$, and
$\inner{u,v}$ denote the inner product of vectors $u$ and $v$. For sequence of $n$ vectors $z_1, \dotsc, z_n$, we use the notation $z{[n]}$ to denote the whole sequence.
For vector $v$, $v^{*m}$ refers to elementwise $m^{\tha}$ power of $v$.
For matrix $C \in \R^{d \times k}$, the $j$-th column is referred by $C_j$ or $c_j$, $j \in [k]$, the $j^{\tha}$ row is referred by $C^{(j)}$ or $c^{(j)}$, $j \in [d]$ and $\|C\|$ denotes the spectral norm of matrix $C$.
 Throughout this paper, $\nabla_x^{(m)}$ denotes the $m^{\tha}$ order derivative operator w.r.t.\ variable $x$. %and notation $\otimes$ represents tensor (outer) product. 	

\paragraph{Tensor:}
% A real \emph{$r$-th order tensor} $T \in \bigotimes_{i=1}^r \R^{d_i}$ is a member of the outer product of Euclidean spaces $\R^{d_i}$, $i \in [r]$.
%For convenience, we restrict to the case where $d_1 = d_2 = \dotsb = d_r = d$, and simply write $T \in \bigotimes^r \R^d$.
%%For vector $v \in \R^d$, we use $v^{\otimes r} := v \otimes v \otimes
%%\dotsb \otimes v \in \bigotimes^r \R^d$ to denote its $r$-th tensor power.
%As is the case for vectors (where $r=1$) and matrices (where $r=2$), we may
%identify a $r$-th order tensor with the $r$-way array of real numbers $[
%T_{i_1,i_2,\dotsc,i_r} \colon i_1,i_2,\dotsc,i_r \in [d] ]$, where
%$T_{i_1,i_2,\dotsc,i_r}$ is the $(i_1,i_2,\dotsc,i_r)$-th coordinate of $T$
%with respect to a canonical basis. For convenience, we limit to third order tensors $(r=3)$ in our analysis, while the results for higher order tensors are also provided.
A real \emph{$m^{\tha}$ order tensor} $T \in \bigotimes^m \R^d$ is a member of the outer product of Euclidean spaces $\R^{d}$.
%For convenience, we limit to third order tensors $(p=3)$ in our analysis. %RG: we have 4 now
The different dimensions of the tensor are referred to as {\em modes}. For instance, for a matrix, the first mode refers to columns and the second mode refers to rows.
%In addition,￼ {\em fibers} are higher order analogues of matrix rows and columns. A fiber is obtained by fixing all but one of the indices of the tensor. %(and is arranged as a column vector).
%For example, for a third order tensor $T\in \R^{d \times d \times d}$, the mode-$1$ fiber is given by $T(:, j, l)$. %, mode-$2$ by $T(i, :, l)$ and mode-$3$ by $T(i, j, :)$.
%Similarly, {\em slices} are obtained by fixing all but two of the indices of the tensor. For example, for the third order tensor $T$, the slices along $3$rd mode are given by $T(:, :, l)$.

\paragraph{Tensor matricization:}
For a third order tensor $T \in \R^{d \times d \times d}$, the matricized version along first mode denoted by $M \in \R^{d \times d^2}$ is defined such that
\begin{equation} \label{eqn:matricization}
T(i,j,l) = M(i,l+(j-1)d), \quad i,j,l \in [d],
\end{equation}
and we use Mat to show matricization, i.e., $M=\text(Mat)(T)$
\paragraph{Tensor Reshaping: }
$T_2=\text{Reshape}(T_1,v_1, \dotsc, v_l)$ means that $T_2$ is a tensor of order $l$ that is made by  reshaping tensor $T_1$ such that the first mode of $T_2$ includes modes of $T_1$ that are shown in $v_1$,
the second mode of $T_2$ includes modes of $T_1$ that are shown in $v_2$ and so on. For example if $T_1$ is a tensor of order $5$,  $T_2=\text{Reshape}(T_1,[1~2],3 ,[4~ 5] )$ is a third order tensor, where its first mode is made by concatenation of modes $1, 2$ of $T_1$ and so on.

\paragraph{Tensor rank:}A $3$rd order tensor $T \in \Rbb^{d \times d \times d}$ is said to be rank-$1$ if it can be written in the form
\begin{equation} \label{eqn:rank-1 tensor}
T= w \cdot a \otimes b\otimes c \Leftrightarrow T(i,j,l) = w \cdot a(i) \cdot b(j) \cdot c(l),
\end{equation}
where $\otimes$  represents the {\em outer product}, and $a, b , c \in \Rbb^d$ are unit vectors.
A tensor $T  \in \Rbb^{d \times d \times d}$ is said to have a CP (Candecomp/Parafac) {\em rank} $k$ if it can be (minimally) written as the sum of $k$ rank-$1$ tensors
\begin{equation}\label{eqn:tensordecomp}
T = \sum_{i\in [k]} w_i a_i \otimes b_i \otimes c_i, \quad w_i \in \Rbb, \ a_i,b_i,c_i \in \Rbb^d.
\end{equation}
Note that $v^{\otimes p}= v \otimes v \otimes v \cdots \otimes v$, where $v$ is repeated $p$ times.

\begin{definition}[Row-wise Kronecker product] \label{KR2}
For matrices $A, B \in \R^{d \times k}$, the {\em Row-wise Kronecker product} $\in \R^{d \times k^2}$ is defined below
\begin{align*}
\left[ \begin{array}{c} a^{(1)} \vspace{-.2em}   \\   \hline  \vspace{-1em}  \\  a^{(2)}  \\ \hline \vspace{-1em}  \\   \vdots  \\ \hline   a^{(k)} \end{array}\right] \odot \left[ \begin{array}{c} b^{(1)} \vspace{-.2em}  \\ \hline \vspace{-1em} \\  b^{(2)} \\ \hline \vspace{-1em}  \\\vdots \\ \hline  b^{(k)} \end{array}\right] = \left[ \begin{array}{c} a^{(1)} \otimes b^{(1)} \vspace{-.2em} \\ \hline \vspace{-1em} \\ a^{(2)}  \otimes b^{(2)} \\ \hline \vspace{-1em} \\ \vdots \\ \hline a^{(k)} \otimes b^{(k)}\end{array}\right],
\end{align*}
where $a^{(i)}, b^{(i)}$ are rows of $A, B$ respectively. Note that our definition is different from usual definition of Khatri-Rao product which is a column-wise Kronecker product (is performed on columns of matrices).
%\hscomment{mention dimension}
\end{definition}

\paragraph{Tensor as multilinear form:} We view a tensor $T \in \Rbb^{d \times d \times d}$ as a multilinear form. %For simplicity let us consider order-3 tensors.
Consider matrices $M_l \in \R^{d\times d_l}, l \in \{1,2,3\}$. Then tensor $T(M_1,M_2,M_3) \in \R^{d_1}\otimes \R^{d_2}\otimes \R^{d_3}$ is defined as
\begin{align} \label{eqn:multilinear form def}
T(M_1,M_2,M_3)_{i_1,i_2,i_3} := \sum_{j_1, j_2,j_3\in[d]} T_{j_1,j_2,j_3} \cdot M_1(j_1, i_1) \cdot M_2(j_2, i_2) \cdot M_3(j_3, i_3).
\end{align}
%In particular, if $u$, $v$ and $w$ are vectors and $T$ is a $3$rd order tensor, then $T(u,v,w)$ is a scalar, $T(I,v,w)$ is a vector, and $T(I, I, w)$ is a matrix.
In particular, for vectors $u,v,w \in \R^d$, we have\,\footnote{Compare with the matrix case where for $M \in \R^{d \times d}$, we have $ M(I,u) = Mu := \sum_{j \in [d]} u_j M(:,j) \in \R^d$.}
\begin{equation} \label{eqn:rank-1 update}
 T(I,v,w) = \sum_{j,l \in [d]} v_j w_l T(:,j,l) \ \in \R^d,
\end{equation}
which is a multilinear combination of the tensor mode-$1$ fibers.
Similarly $T(u,v,w) \in \R$ is a multilinear combination of the tensor entries,  and $T(I, I, w) \in \R^{d \times d}$ is a linear combination of the tensor slices.

\paragraph{Derivative:}
For function $g(x): \R^d \rightarrow \R$ with vector input $x \in \R^d$, the $m$-th order derivative w.r.t.\ variable $x$ is denoted by $\nabla_x^{(m)} g(x) \in \bigotimes^{m} \R^d$ (which is a $m$-th order tensor) such that
\begin{equation} % \label{eqn:derivativedef}
\left[ \nabla_x^{(m)} g(x) \right]_{i_1,\dotsc,i_m} := \frac{\partial g(x)}{\partial x_{i_1} \partial x_{i_2} \dotsb \partial x_{i_m}}, \quad i_1,\dotsc,i_m \in [d].
\end{equation}
%In addition, the $m$-th order derivative is denoted by $\nabla_x^{(m)} F(x) \in \bigotimes^{r+m} \R^d$.
When it is clear from the context, we drop the subscript $x$ and write the derivative as $\nabla^{(m)} g(x)$.

\paragraph{Derivative of product of two functions}
 We frequently use the following gradient rule.
\begin{lemma} [Product rule for gradient~\citep{janzamin2014matrix}] \label{lem:prodrule}
For tensor-valued functions
$F(x) : \R^n \rightarrow\bigotimes^{p_1} \R^n,
G(x) : \R^n \rightarrow\bigotimes^{p_2} \R^n $, we have
\[
\nabla_x (F(x) \otimes G(x)) = (\nabla_x F(x) \otimes G(x))^{\langle \pi \rangle} + F \otimes \nabla_x G(x),
\]
where the notation $^{\langle \pi \rangle}$ denotes permutation of modes of the tensor for permutation vector $\pi = [1,2,\dotsc,p_1,p_1+2,p_1+3,\dotsc,p_1+p_2+1,p_1+1]$. This means that the $(p_1+1)^{\tha}$ mode is moved to the last mode.
\end{lemma}

\section{Proof of Theorems~\ref{lemma:RNN-Q},~\ref{lemma:BRNN}} \label{appendix:proofs}

%\subsection{GLOREE for general polynomials} \label{appendix:RNN-Alg}

\subsection{Proof Theorem~\ref{lemma:RNN-Q}} \label{appendix:RNN-Q}
%\begin{proof}
The underlying idea behind the proof comes from Theorem~\ref{thm:steins_higher}. By Theorem~\ref{thm:steins_higher} we have that
\begin{align*}
\Ebb \left[ y_t \otimes \Sc_2(x_t) \right] = \Ebb\left[\nabla^2_{x_t} \Ebb \left[  y_t | x_t \right]\right].
\end{align*}
In order to show the derivative form more easily, let us look at derivative of each entry $i \in [d_y]$ of the vector $y_t$.
\begin{align*}
\Ebb \left[  (y_t)_i | x_t \right] &= \inner{(A_2)_{(i)}, \left(A_1 x_t+ U h_{t-1} \right)^{*2} } = \sum_{j \in d_h} (A_2)_{ji} \left(\inner{A_1^{(j)},x_t}+\inner{U^{(j)},h_{t-1}}\right)^2,  \\
\Ebb[\nabla^2_{x_t} \Ebb[(y_t)_i| x_t]] &= \Ebb\left[\nabla^2_{x_t} \left(\sum_{j \in d_h} (A_2)_{ji} \left(\inner{A_1^{(j)},x_t}+\inner{U^{(j)},h_{t-1}}\right)^2\right)\right] \\
& = \Ebb\left[ 2\sum_{j \in d_h} (A_2)_{ji} \nabla_{x_t}  \left(\inner{A_1^{(j)},x_t}+\inner{U^{(j)},h_{t-1}}\right)A_1^{(j)}\right] \\
& =  2\sum_{j \in d_h} (A_2)_{ji} A_1^{(j)} \otimes A_1^{(j)},\\
\Ebb[\nabla^2_{x_t} \Ebb \left[  y_t | x_t \right]] &=2\sum_{j \in d_h} A_2^{(j)} \otimes A_1^{(j)} \otimes A_1^{(j)},
\end{align*}
and hence the form follows.
By Theorem~\ref{thm:steins_higher} we have that
\begin{align*}
\Ebb \left[ y_t \otimes \Sc_4(x_{t-1}) \right] = \Ebb\left[ \nabla^4_{x_{t-1}}\Ebb \left[ y_t | x_{t-1} \right]\right]
\end{align*}
In order to show the derivative form more easily, let us look at each entry $i \in [d_y]$ of the vector $y_t$.
\begin{align*}
(h_{t})_k &= \left(\inner{A_1^{(k)}, x_t}+\sum_{l \in [d_h]} U_{kl} \left( \inner{A_1^{(l)}, x_{t-1}}+\inner{U^{(l)}, h_{t-2}} \right)^2 \right)^2 \\
\Ebb[(y_t)_i|x_{t-1}] &= \Ebb\left[\sum_{k \in [d_h]} (A_2)_{ki} \left(\inner{A_1^{(k)}, x_t}+\sum_{l \in [d_h]} U_{kl} \left( \inner{A_1^{(l)}, x_{t-1}}+\inner{U^{(l)}, h_{t-2}} \right)^2 \right)^2\right].
\end{align*}
The form follows directly using the derivative rule as in Lemma~\ref{lem:prodrule}.
\begin{align*}
T= \Ebb\left[\nabla^4_{x_{t-1}} \Ebb[ y_t | x_{t-1}] \right]&=2\sum_{j \in d_h} A_2^{(j)} \otimes \sum_{k \in d_h} U_{jk}A_1^{(k)} \otimes A_1^{(k)} \otimes  \sum_{m \in d_h} U_{jm}A_1^{(m)} \otimes A_1^{(m)},
\end{align*}
Now when we reshape the above tensor $T$ to $\text{Reshape}(\Ebb[ \nabla^4_{x_{t-1}} \Ebb[y_t| x_{t-1}], [1~2], [3~ 4]])$ we have the form $\sum_{j \in d_h} A_2^{(j)} \otimes [U(A_1 \odot A_1)]^{(j)} \otimes [U(A_1 \odot A_1)]^{(j)}.$
%\end{proof}
\paragraph{Remark: Difference from IID case: }Note that Lemma 2 is different from Theorem 1. Lemma 2 is specific to RNNs while Theorem 1 is from [Janzamin et al 2014] for IID samples. The score function in the two results are different; Note that in Lemma 2 score function $S_m( x[n],t )$ in Equation $(3)$ is defined as partial derivative (of order $m$) of joint pdf $p(x_1, \dotsc, x_t)$ w.r.t. $x_t$ which is a m-th order tensor. When this form is utilized in Stein’s form in Lemma 2, we obtain partial derivatives of $y_t$ w.r.t. $x_t$ which is a function of $x_1,\dotsc, x_t$ in expectation. Note that the expectation is w.r.t. all variables $x_1,\dotsc, x_t,$ and thus, the dependence to $h_{t-1}$ is also averaged out since it is a function of $x_1,\dotsc,x_{t-1}$. To provide more steps: let 
$G(x_1,\dotsc, x_t):= E[y_t| x_1,\dotsc, x_t]$.
Using law of total expectation and Theorem 1, we have
\begin{align*}
\Ebb[y_t \otimes  S_m(x[n],t)]=\Ebb[\nabla_{x_t}^{(m)} G(x_1,\dotsc, x_t)].
\end{align*}
Since $G(x_1, \dotsc, x_t)=A_2^T poly( A_1 x_t + U h_{t-1})$ where $h_{t-1}$ is only a function of $x_1,\dotsc, x_{t-1}$, the result follows. Again note that the partial derivative on the RHS is only w.r.t. $x_t$, while the expectation is for all $x_1, \dotsc, x_t$. The crucial point is $G$ is a function of $x_1,\dotsc, x_t$ and not just $x_t$. It is the use of partial derivatives that allows us to carry out this operation. This is also a novel contribution of this paper and does not follow directly from score function result in [Janzamin et al 2014]. 
\subsection{Proof of Theorem~\ref{lemma:BRNN}} \label{appendix:proofsBRNN}
\begin{align*}
\Ebb[y_t | h_t, z_t]  = A_2^\top \left[\begin{array}{c} h_t \\ z_t .
\end{array} \right]
\end{align*}
Hence,
\begin{align*}
T & = \Ebb\left[\nabla_{x_t}^2 \Ebb[y_t | h_t, z_t]\right] = \Ebb\left[A_2^\top \left[\begin{array}{c} \nabla_{x_t}^2 h_t \\ \nabla_{x_t}^2 z_t.
\end{array} \right]\right] \\
&=A_2 ^\top \left[\begin{array}{c} \sum_{i \in d_h} e_i \otimes (A_1)^{(i)} \otimes (A_1)^{(i)} \\ \sum_{i \in d_h} e_i \otimes (B_1)^{(i)} \otimes (B_1)^{(i)}.
\end{array} \right]
\end{align*}
The second Equation is direct result of Lemma~\ref{lemma:RNN-Q}. Therefore, if we decompose the above tensor, the first mode yields the matrix $A_2$. Next we remove the effect of $A_2$ by multiplying its inverse to the first mode of the moment tensor $T$.
By the above Equations, we readily see that
\begin{align*}
T( (A_2)^{-1}, I, I) = \left[\begin{array}{c} \sum_{i \in d_h} e_i \otimes (A_1)^{(i)} \otimes (A_1)^{(i)} \\ \sum_{i \in d_h} e_i \otimes (B_1)^{(i)} \otimes (B_1)^{(i)}.
\end{array} \right]
\end{align*}
This means $T( (A_2)^{-1}, I, I) = \sum_{i \in d_h} e_i \otimes c_i \otimes c_i$, where $c_i = \left[\begin{array}{c} (A_1)^{(i)} \\ (B_1)^{(i)}
\end{array} \right]. $
Hence, Algorithm~\ref{algo:BRNN} correctly recovers $A_2, A_1, B_1$.
Recovery of $U, V$ directly follows Lemma~\ref{lemma:RNN-G}.

%\section{GLOREE-B for BRNN with quadratic activation function} \label{appendix:BRNNg1}
%In Algorithm~\ref{algo:BRNN}, we show the algorithm for training BRNN when the activation functions are quadratic.
%\input{BRNN-alg}

\section{GLOREE for General Polynomial Activation Functions}
\subsection{GLOREE for IO-RNN with polynomial activation functions} \label{appendix:RNN-Alg}
%\label{sec:higherorder}
\begin{algorithm}[b]
\caption{GLOREE (Guaranteed Learning Of Recurrent nEural nEtworks) for vector input}
\label{algo:mainRNN_poly}
\begin{algorithmic}[1]
\renewcommand{\algorithmicrequire}{\textbf{input}}
\renewcommand{\algorithmicensure}{\textbf{output}}
\REQUIRE (a) Labeled samples $\{(x_i,y_i): i \in [n]\}$ from IO-RNN model in Figure~\ref{fig:RNN}(b), polynomial order $l$ for activation function.

%\IF {$l = 1$}
%\STATE
%{$\forall y_i, i \in [n]$ Replace $y_i$ with ${y_i}^{*2}$}
%\ENDIF
\STATE Compute $2$nd-order score function $\Pc_2(x[n],i)$ of the input sequence as in Equation~\eqref{eqn:sf-mc}.
\STATE Compute $\widehat{T}:= \widehat{\Ebb}\left[ y_i \otimes \Pc_2(x[n], i) \right]$.The empirical average is over a single sequence.
% $\widehat{D}:= \widehat{\Ebb}\left[ y_i \otimes \Pc_1(x[n], i) \right]$.  and $\widehat{D}$ will be used for whitening (Procedure~\ref{algo:whitening}).

%\STATE Compute $\widehat{T}:= \frac{1}{n} \sum_{i \in [n]} y_i \cdot \Pc_2(x_i)$. %, Empirical estimate of
\STATE $\lbrace \hat{w}, \hat{R}_1, \hat{R}_2,\hat{R}_3  \rbrace =\text{tensor decomposition}(\widehat{T})$; see Appendix~\ref{sec:tensor}. \label{line:TensorDecomposition}
\STATE $\hat{A}_2=\hat{R}_1,   \hat{R}_1= \left(  \hat{R}_2 + \hat{R}_3 \right)/2 $.
%\STATE $\hat{A}_1$ is the matrix whose columns are $\lbrace u_j \rbrace_{j \in [k]}$.

%\STATE $q=\max \lbrace 2, l^2 \rbrace$
\STATE Compute $ l^2$th-order score function $\Pc_{l^2}(x[n], i)$ of the input sequence  as in Equation~\eqref{eqn:sf-mc}.

\STATE Compute $\widehat{T}=\hat{\Ebb} \left[y_t\otimes \text{Reshape}(\Sc_{l^2}(x[n],{t-1}), [1 ~\dotsc~ l],\dotsc, [l^2-l+1~ \dotsc ~l^2] \right]$.
%\tcr{\STATE Contract the moment-tensor to a $3$-rd order tensor as $\widehat{T}_2=T(I, I, I, \theta, \dotsc, \theta)$ with some random vector $\theta$.}
%\STATE Compute $\widehat{T}:= \frac{1}{n} \sum_{i \in [n]} y_i \cdot \Pc_2(x_i)$. %, Empirical estimate of
\STATE $\lbrace \hat{w}, \hat{R_1}, \hat{R_2}, \hat{R_{3}}  \rbrace =\text{tensor decomposition}(\widehat{T})$; using sketching~\citep{wang2015fast}. \label{line:Bsketching}
\STATE $\tl{R}=\left(  \hat{R}_2 + \hat{R}_3 \right)/2$.
\STATE $\hat{U}=\tl{R}\left[\hat{A}_1 \odot \hat{A}_1  \right]^{\dagger}$, row-wise Kronecker product $\odot$ is defined in Definition~\ref{KR2}.

\RETURN $\hat{A}_1, \hat{A}_2, \hat{U}$.

\end{algorithmic}
\end{algorithm}

Here,  we consider an RNN with polynomial activation function of order $l \geq 2$, i.e.,
\begin{align} \label{eq:RNN}
\Ebb[ y_t| h_t] = A_2^\top h_t, \quad
h_t = \poly_l(A_1 x_{t}+ U h_{t-1}).
\end{align}
We have the following properties.
\begin{theorem}[Learning parameters of RNN for general polynomial activation function] \label{lemma:RNN-G}The following is true:
\begin{align}
\Ebb \left[ y_t \otimes \Sc_2(x[n], t) \right] &= \sum_{i \in d_h} \mu_i A_2^{(i)}  \otimes A_1^{(i)} \otimes A_1^{(i)}.
\end{align}\label{eqn:moment1}
Hence, we can recover $A_1, A_2$ via tensor decomposition assuming that they are full row rank.
%\end{lemma}

In order to learn $U$ we form the tensor $\Ebb \left[y_t \otimes \Sc_{l^2}(x_{t-1}) \right]$.
Then we have
%\begin{lemma}
\begin{align}\label{eqn:moment2}
\Ebb \left[ y_t\otimes \text{Reshape}(\Sc_{l^2}(x[n], {t-1}),1, [1 ~\dotsc~ l],\dotsc, [l^2-l+1~ \dotsc ~l^2] \right] =  \sum_{i \in d_h} A_2^{(i)}  \otimes  \left[\left[U (A_1^{\odot l}) \right]^{(i)}\right]^{\otimes l}.
\end{align}
Hence, we can recover $U(A_1^{\odot l})$ via tensor decomposition under full row rank assumption. Since $A_1$ is previously recovered, $U$ can be recovered. Thus, Algorithm~\ref{algo:mainRNN_poly}(GLOREE) consistently recovers the parameters of IO-RNN with polynomial activations.
\end{theorem}

\paragraph{Remark on form of the cross-moment tensor:} The cross-moment tensor in Equation~\eqref{eqn:moment2}, is a tensor of order $l+1$, where modes $2, \dotsc, l+1$ are similar, i.e., they all correspond to rows of the matrix $U (A_1^{\odot l})= U(A_1 \odot A_1 \dotsc \odot  A_1)$, where $A_1$ has gone through row-wise Kronecker product $l$ times~\footnote{Since row-wise Kronecker product (as defined in notations) does not change the number of rows, this matrix multiplication is valid.}. This is a direct extension of the form in Theorem~\ref{lemma:RNN-Q} from $l=2$ to any $l \geq 2$.

\paragraph{Remark on coefficients $\mu_i$: }For the   cross-moment tensor in \eqref{eqn:moment1}, the coefficients $\mu_i$ are the expected values of  derivatives of activation function. More concretely, if activation is a polynomial of degree $l$, we have that  $\mu_i =\Ebb \left[ \poly_{l-2}\left( \inner{A_1^{(i)}, x_t}+\inner{U^{(i)},h_{t-1}} \right)    \right]$, where $\poly_{l-2}$ denotes a polynomial of degree $l-2$. Similarly, the coefficients of the tensor decomposition in \eqref{eqn:moment2} correspond to expectations over derivatives of (recursive) activation functions. We assume that these coefficients are non-zero in order to recover the weight matrices.

\paragraph{Remark on tensor decomposition via sketching: } Consider line~\ref{line:sketching} in Algorithm~\ref{algo:mainRNN_poly} and line~\ref{line:Bsketching} in Algorithm~\ref{algo:BRNN-general}. Here we are decomposing a tensor of order $l+1$.  In order to perform this with efficient computational complexity, we can use tensor sketching proposed by~\citet{wang2015fast}. They do not form the moment tensor explicitly and directly compute tensor sketches from data. This avoids the exponential blowup in computation, i.e., it reduces the computational complexity from $m^{l+1}$ to $(m+m\log m)n$, where $m$ is the sketch length and $n$ denotes the number of samples. As expected, there is a trade off between the sketch length and the error in recovering the tensor components. For details, see~\citep{wang2015fast}.
%
%\paragraph{Remark: Recovering $U$: }  Note that after the first step, the only remaining parameter is the matrix $U \in \mathbb{R}^{d_h \times d_h}$. Hence we have greatly  reduced the problem dimension. Therefore, at this point we can proceed in two ways, we can either use gradient descent to search for $U$ or proceed with the moment-based approach. In this paper we have used the latter approach.

\bprf
By Theorem~\ref{thm:steins_higher}, we have that
\begin{align*}
&\Ebb \left[ y_t \otimes \Sc_2(x[n], t) \right] = \Ebb\left[\nabla^2_{x_t} \Ebb \left[ y_t | x_t \right]\right],\\
&\Ebb \left[ y_t\otimes \text{Reshape}(\Sc_{l^2}(x[n], {t-1}),[1 ~\dotsc~ l],\dotsc, [l^2-l+1~ \dotsc ~l^2])\right] \\ &= \Ebb \left[ \text{Reshape}(\nabla^{l^2}_{x_t} y_t ,1, [2 ~\dotsc~ l+1],\dotsc, [l^2~ \dotsc ~l^2+1]) \right].
\end{align*}
The form follows directly using the derivative rule as in Lemma~\ref{lem:prodrule}.

\eprf

%In the second part of the Algorithm, we form a tensor of order $l+1$ in practice  the activation function is usually a low-order polynomial and $l$ is small. Even otherwise, we have already retrieved estimates of $A_1, A_2$ and the problem has shrunk noticeably.
%Note that in order to recover $U$, we can either decompose a tensor of order $l+1$ or contract the tensor to a third-order tensor and then decompose the contracted tensor. For simplicity, we show the latter in our algorithm.

Thus, we provide an efficient framework for recovering all the weight matrices of an input-output recurrent neural network using tensor decomposition methods.

\subsection{GLOREE-B for BRNN with general polynomial activation function} \label{appendix:BRNNg}

In Algorithm~\ref{algo:BRNN-general}, we show the complete algorithm for training BRNN when the activation functions are polynomials of order $l$. The analysis directly follows from analysis of BRNNs with quadratic activation functions and the extension to $l \geq 2$ is similar to extension of IO-RNNs with quadratic activation functions to general polynomials of order $l \geq 2$.

\begin{algorithm}[t]
\caption{GLOREE-B (Guaranteed Learning Of Recurrent nEural nEtworks-Bidirection case) for general activation function}
\label{algo:BRNN-general}
\begin{algorithmic}[1]
\renewcommand{\algorithmicrequire}{\textbf{input}}
\renewcommand{\algorithmicensure}{\textbf{output}}
\REQUIRE Labeled samples $\{(x_i,y_i): i \in [n]\}$, polynomial order $l_h$ for activation function in the forward direction,  polynomial order $l_z$ for activation function in the backward direction.

\REQUIRE $2$nd-order score function $\Pc_2(x[n], [n])$ of the input $x$; see Equation~\eqref{eqn:highorder} for the definition.

\STATE Compute $\widehat{T}:= \hat{\Ebb} [y_i \otimes \Pc_2(x[n], i)]$.
%, $\widehat{D}:= \hat{\Ebb} [y_i \otimes \Pc_1(x[n], i)]$. $\widehat{D}$ will be used for whitening (Procedure~\ref{algo:whitening}).
\STATE $\lbrace (\hat{w}, \hat{R}_1, \hat{R}_2,\hat{R}_3  \rbrace=\text{tensor decomposition}(\widehat{T})$; see Appendix~\ref{sec:tensor}.

\STATE $\hat{A}_2 = \hat{R}_1 $.
\STATE  Compute $\tl{T}=\widehat{T}(((\hat{A}_2)^\top)^{-1}, I, I)$. For definition of multilinear form see Section~\ref{sec:notation}.
\STATE $\lbrace (\hat{w}, \hat{R}_1, \hat{R}_2,\hat{R}_3  \rbrace=\text{tensor decomposition}(\tl{T})$;
\STATE $   \hat{C}=(\hat{R}_2+\hat{R}_3)/2 $.
\STATE $\hat{C}=\left[\begin{array}{c}
\hat{A}_1 \\
\hat{B}_1
\end{array}\right] $.

\STATE Compute $ l^2$th-order score function $\Pc_{ l^2}(x[n],{t-1})$ of the input sequence  as in Equation~\eqref{eqn:sf-mc}. \label{line:beginG}

\STATE Compute $\widehat{T}=\hat{\Ebb} \left[y_t\otimes \text{Reshape}(\Sc_{l^2}(x[n], {t-1}),1, [1 ~\dotsc~ l],\dotsc, [l^2-l+1~ \dotsc ~l^2] \right]$.
%\tcr{\STATE Contract the moment-tensor to a $3$-rd order tensor as $\widehat{T}_2=T(I, I, I, \theta, \dotsc, \theta)$ with some random vector $\theta$.}
%\STATE Compute $\widehat{T}:= \frac{1}{n} \sum_{i \in [n]} y_i \cdot \Pc_2(x_i)$. %, Empirical estimate of
\STATE $\lbrace \hat{w}, \hat{R_1}, \hat{R_2}, \hat{R_{3}}  \rbrace =\text{tensor decomposition}(\widehat{T})$; using sketching~\citep{wang2015fast}. \label{line:sketching}
\STATE $\tl{R}=\left(  \hat{R}_2 + \hat{R}_3 \right)/2$.
\STATE $\hat{U}=\tl{R}\left[\hat{A}_1 \odot \hat{A}_1 \right]^{\dagger}$, row-wise Kronecker product $\odot$ is defined in Definition~\ref{KR2}. \label{line:endG}
%\STATE Compute $\widehat{T}:= \frac{1}{n} \sum_{i \in [n]} y_i \cdot \Pc_2(x_i)$. %, Empirical estimate of
%\STATE Compute $\widehat{T}=\hat{\Ebb} \left[y_t\otimes \text{Reshape}(\Sc_{l^2}(x[n],{t+1}),1, [1 ~\dotsc~ l],\dotsc, [l^2-l+1~ \dotsc ~l^2] \right]$.
%\STATE Contract the moment-tensor to a $3$-rd order tensor as $\widehat{T}_2=T(I, I, I, \theta, \dotsc, \theta)$ where $\theta \sim \mathcal{N}(0,I_{d})$ is a random standard Gaussian vector.
%%\STATE Compute $\widehat{T}:= \frac{1}{n} \sum_{i \in [n]} y_i \cdot \Pc_2(x_i)$. %, Empirical estimate of
%\STATE $\lbrace \hat{w}, \hat{R_1}, \hat{R_2}, \hat{R_{3}}  \rbrace =\text{tensor decomposition}(\widehat{T}_2)$; see Algorithm~\ref{algo:main} in the Appendix.
%\STATE $\tl{R}=\left(  \hat{R}_2 + \hat{R}_3 \right)/2$.
%\STATE $\hat{V}=\tl{R}\left[\hat{B}_1 \odot \hat{B}_1 \right]^{-1}$, Khatri-Rao product $\odot$ is defined in Definition~\ref{KR}.
\STATE Repeat lines~\eqref{line:beginQ}-\eqref{line:endQ} with $\Pc_{4}(x[n],{t+1})$ instead of $\Pc_{4}(x[n],{t-1})$ to recover $\hat{V}$.

%\STATE Compute $u_1$ and $v_1$ as the top left and right singular vectors of  $T(I,I,\theta) \in \R^{d \times d}$.
%\STATE $\ha^{(0)} \leftarrow u_1$, $\hb^{(0)} \leftarrow v_1$.
%\STATE Initialize $\hc^{(0)}$ by update formula in \eqref{eqn:asymmetric power update}.
\RETURN $\hat{A}_1, \hat{A}_2, \hat{B}_1, \hat{U}, \hat{V}$.

\end{algorithmic}
\end{algorithm}
\section{Sample complexity analysis Proofs} \label{sec:samplecomp}
In this Section, we provide the proofs corresponding to sample complexity analysis of Section~\ref{main:sample} in the paper. We also elaborate on the Assumptions 2(d) and 2(e) and discuss that in fact we need weaker assumptions. 

We now analyze the sample complexity for GLOREE. We first start with the concentration bound for the moment tensor and then use analysis of tensor decomposition to show that our method has a sample complexity that is a polynomial of the model parameters.
%
%
%We first derive concentration bounds for the empirical moment tensors. This is a bit more involved since we have a non-i.i.d. sequence. Below we elaborate on our assumptions ad why they are needed. 
%%Moreover, since we assume a  polynomial activation function, the hidden state $h_t$ can grow in an unbounded manner. To avoid this, without loss of generality, we assume $\| x_i \|_2 <1,~~ \forall i \in [n]$ (with high probability). We also need the three additional assumptions mentioned below. %Our analysis shows that we also require $\| A_1 + U \|\leq 1$.
%
%\paragraph{Assumptions: }\begin{enumerate} \item \textbf{Ensure bounded hidden variables :} since we assume a  polynomial activation function, the hidden state $h_t$ can grow in an unbounded manner. To avoid this we need the following: \begin{enumerate} \item Without loss of generality, we assume that the input sequence is bounded by $1$ with high probability, $\| x_i \| <1~~ \forall i \in [n]$ \item Assume that $\| A_1 \| + \| U \| \leq 1$.
%\end{enumerate}
%\item \textbf{Ensure Concentration of moment tensor: }
%\begin{enumerate}
%\item The input sequence is a polynomial ergodic Markov chain.
%\item If the activation function is a polynomial of order $l$, we need $\| U\| \leq 1/l$.
%\item $\Vert\Pc_2(x[n], i)\Vert$ is a bounded value.
%\item The input sequence is a first order Markov chain.
%\item $\left\lbrace\|\Pc_3(x[n], i) - \Pc_2(x[n], i) \otimes \Pc_1(x[n], i)\|\right\rbrace$ is bounded.
%\end{enumerate} \end{enumerate}

\paragraph{Concentration bounds for functions over Markov chains: }Our input sequence $x[n]$ is a geometrically ergodic Markov chain. We can think of  the empirical moment $\hat{\Ebb}[y_t\otimes \Sc_m(x[n], t)]$ as functions  over the samples $x[n]$ of the Markov chain. Note that this assumes $h[n]$ and $y[n]$ as deterministic functions of $x[n]$, and our analysis can be extended when there is additional randomness.~\citet{kontorovich2014uniform} provide the result for scalar functions and this is an extension of that result to matrix-valued functions.

We now recap concentration bounds for general functions on Markov chains. For any ergodic Markov chain with the stationary distribution $\omega$, denote $f_{1 \rightarrow t}(x_t|x_1)$ as state distribution given initial state $x_1$.  The inverse mixing time is defined as follows
\begin{align*}
\rho_{\text{mix}}(t)= \sup_{x_1}\Vert f_{1 \rightarrow t}(x_t|x_1)-\omega\Vert.
\end{align*} ~\citet{kontorovich2014uniform} show that
\begin{align*}
\rho_{\text{mix}}(t) \leq G\theta^{t-1},
\end{align*}
where $1 \leq G < \infty$ is geometric ergodicity and $0 \leq \theta < 1$ is the contraction coefficient of the Markov chain.

In the IO-RNN (and BRNN) model, the output is a nonlinear function of the input. Hence,  the next step is to deal with this non-linearity. \citet{kontorovich2014uniform} analyze the mixing of a (scalar) nonlinear function through its Lipschitz property. In order to analyze how the empirical moment tensor concentrates, we define the Lipschitz constant for matrix valued functions.

\begin{definition}[Lipschitz constant for a matrix-valued function of a sequence]
A matrix-valued function $\Phi: \Rbb^{n d_x} \rightarrow \mathbb{R}^{d_1 \times d_2}$ is $c$-Lipschitz  with respect to the spectral norm if
\begin{align*}
\sup_{x[n], \tl{x}[n]} \frac{\Vert \Phi(x[n])-\Phi(\tl{x}[n]) \Vert}{\Vert x[n]-\tl{x}[n]  \Vert_2 } \leq c,
\end{align*}where   $x[n], \tl{x}[n]$ are any two possible sequences of observations. Here $\|\cdot \|$ denotes the spectral norm  and $\Rbb^{n d_x}$ is the state space for a sequence of $n$ observations $x[n]$.
\end{definition}

\paragraph{Concentration of empirical moments of IO-RNN and BRNN: }In order to ensure that the empirical moment tensor has a bounded Lipschitz constant, we  need Assumptions 1(a)-1(b) and 2(a)-2(e). Then, we have that
%We prove the Lipschitz property in Lemma~\ref{lemma:LipschitzRNN} in the Appendix~\ref{sec:Appendix-concentration}.

%Under the above assumptions, we have that
\begin{lemma}[Lipschitz property for the Empirical Moment Tensor] \label{lemma:LipschitzRNNcomplete}
For the IO-RNN discussed in~\eqref{eq:RNN}, if the Assumptions 1(a)-1(b), 2(a)-2(e) hold, the matricized tensor $\text{Mat}\left(\hat{\Ebb}[y_t \otimes \Pc_2(x[n], t)]\right)$ is a function of the input sequence with Lipschitz constant
\begin{align}
\label{eqn:lipschitz-const}
c \leq \frac{1}{n}\Vert A_2\Vert  \left[\frac{\Vert A_1 \Vert }{1-l \Vert U \Vert } \Vert\Pc_2(x[n], t)\Vert+ 3 \gamma \right]
%\tcr{c \leq \frac{1}{n}\frac{\Vert A_2\Vert \Vert A_1 \Vert }{1-l \Vert U \Vert } \underset{i \in [n]}{\max} ~\Vert  \Pc_2(x[n], i)\Vert+\Vert A_2\Vert \Vert ~\underset{i \in [n]}{\max} ~\left\lbrace\|\Pc_3(x[n], i) - \Pc_2(x[n], i) \otimes \Pc_1(x[n], i)\|\right\rbrace.}
\end{align}
\end{lemma}
For proof, see Appendix~\ref{appendix:Liptensor}.

Given this Lipschitz constant, we can now apply the following concentration bound.

Now that we have proved the Lipschitz property for the cross-moment tensor, we can prove the concentration bound for the IO-RNN.

\begin{theorem}[Concentration bound for RNN] \label{thm:concentation}
For the IO-RNN discussed in~\eqref{eq:RNN}, let $z[n]$ be the sequence of matricized empirical moment tensors $\text{Mat}\left(\hat{\Ebb}[y_i \otimes \Pc_2(x[n], i)]\right)$ for $i \in [n]$. Then,
\begin{align*}
 \| z - \Ebb(z) \|  &\leq G\frac{1+\frac{1}{\sqrt{8}cn^{1.5}}}{1-\theta}\sqrt{8c^2n\log\left(\frac{d_y+d_x^2}{\delta}\right)},
\end{align*}
 with probability at least $1-\delta$, where $\Ebb(z)$ is expectation over samples of Markov chain when the initial distribution is the stationary distribution and $c$ is specified in Equation~\eqref{eqn:lipschitz-const}.

\end{theorem}

For proof, see Appendix~\ref{sec:Appendix-concentration}.

\subsection{Proof of lemma~\ref{lemma:LipschitzRNNcomplete}:  Lipschitz property for the Empirical Moment Tensor} \label{appendix:Liptensor}
In order to prove lemma~\ref{lemma:LipschitzRNNcomplete}, we first need to show that the matricized cross-moment tensor is a Lipschitz function of the input sequence and find the Lipschitz constant.

We first show that the output function is a Lipschitz function of the input sequence and find the Lipschitz constant. In order to prove this, we need the above assumptions to ensure a bounded hidden state and a bounded output sequence. Then, we have that
%Without loss of generality, we assume that the input sequence is bounded by $1$ with high probability, $\Vert x_i \Vert <1~~ \forall i \in [n]$ (note that this constant in bound of the input sequence is transferable to a bound on norm of $A_1$) and $\Vert A_1\Vert + \Vert U \Vert \leq 1$ (This constant is transferable to the bound on spectral norm of $U$. These two assumptions ensure that $\forall t \in [n], \Vert h_t \Vert \leq 1$.
%Then, if the activation function is a polynomial of order $l$, we need $\Vert U\Vert \leq 1/l$.  We have that

\begin{lemma}[Lipschitz property for the Output of IO-RNN] \label{lemma:LipschitzRNN}
For the IO-RNN discussed in~\eqref{eq:RNN}, if the above assumptions hold, then the output is a Lipschitz function of the input sequence with Lipschitz constant $\frac{1}{n}\frac{\Vert A_2\Vert \Vert A_1 \Vert }{1-l \Vert U \Vert }$ w.r.t. $\ell_2$ metric. 
\end{lemma}

%\begin{proof}[Lipschitz property for the Output of IO-RNN]
\bprf
 This follows directly from the definition. In order to find the Lipschitz constant, we need to sum over all possible changes in the input sequence~\citep{kontorovich2008concentration}. Therefore, we bound derivative of the function w.r.t. each input entry and then take the average the results to provide an upper bound on the Lipschitz constant. With the above assumptions, it is straightforward to show that
 $$ \Vert \nabla_{x_i} y_t \Vert \leq l^{t-i+1} \Vert A_2\Vert \Vert A_1 \Vert \Vert U \Vert^{t-i}.$$

Taking the average of this geometric series for $t \in [n]$ and large sample sequence, we get $\frac{1}{n}\frac{\Vert A_2\Vert \Vert A_1 \Vert }{1-l \Vert U \Vert }$ as the Lipschitz constant.
\eprf 
 %\end{proof}

Next we want to find the Lipschitz constant for the matricized tensor $T=\Ebb[y_t \otimes \Pc_2(x[n], t)]$ which is a function of the input sequence. We use the Assumptions 1(a)-1(c) and 2(a)-2(e) from Section~\ref{main:sample}.

Considering the rule for derivative of product of two functions as in Lemma~\ref{lem:prodrule} we have that
\begin{align*}
\nabla_{x_i} \left[y_t \otimes \Pc_2(x[n], t) \right] = \Pc_2(x[n], t) \otimes \nabla_{x_i} y_t+ y_t \otimes \nabla_{x_i}\Pc_2(x[n], t)
\end{align*}
%Note that by definition of score function in~\eqref{eqn:highorderexpand}, we have that
%\begin{align*}
%\nabla_{x_i}\Pc_2(x[n], i)=\Pc_3(x[n], i)-\Pc_2(x[n], i)\otimes \Pc_1(x[n], i).
%\end{align*}
Hence,
\begin{align*}
\|\nabla_{x_i} \left[y_t \otimes \Pc_2(x[n], t) \right] \| &=\| \Pc_2(x[n], t) \otimes \nabla_{x_i} y_t  + y_t \otimes \nabla_{x_i}\Pc_2(x[n], t)\| \\
& \leq \Vert\Pc_2(x[n], t)\Vert \Vert  \nabla_{x_i} y_t\Vert +\Vert A_2 \Vert\| \nabla_{x_i}\Pc_2(x[n], t) \| ,
\end{align*} 
 we have that
 \begin{align*}
 &\frac{1}{n}\sum_{i \in [n]} \|\nabla_{x_i} \left[y_t \otimes \Pc_2(x[n], t) \right] \\ &~~\leq
\frac{1}{n}\frac{\Vert A_2\Vert \Vert A_1 \Vert \Vert\Pc_2(x[n], t)\Vert }{1-l \Vert U \Vert }+ \frac{1}{n} \Vert A_2\Vert \left\lbrace \| \nabla_{x_{t-1}}\Pc_2(x[n], t) \| + \| \nabla_{x_{t}}\Pc_2(x[n], t) \| + \| \nabla_{x_{t+1}}\Pc_2(x[n], t) \| \right\rbrace \\ 
& ~~\leq \frac{1}{n}\frac{\Vert A_2\Vert \Vert A_1 \Vert \Vert\Pc_2(x[n], t)\Vert }{1-l \Vert U \Vert }+ \frac{1}{n} \Vert A_2\Vert 3 \gamma
 \end{align*}
 the last inequality follows from definition of first order Markov chain.
 and assuming that $\Vert\Pc_2(x[n], i)\Vert$ is bounded and the each of the above derivatives is bounded by some value $\gamma$.
 and we conclude that $\text{Mat}\left(\Ebb[y_i \otimes \Pc_2(x[n], i)]\right)$ is Lipschitz with Lipschitz constant 
\begin{align*}
c = \frac{1}{n}\Vert A_2\Vert  \left[\frac{\Vert A_1 \Vert }{1-l \Vert U \Vert } \Vert\Pc_2(x[n], it)\Vert+  3\gamma \right]
\end{align*}
%\end{proof}

Now that we have proved the Lipschitz property for the cross-moment tensor, we can prove the concentration bound for the IO-RNN.

%\begin{theorem}[Concentration bound for RNN] \label{thm:concentation}
%For the IO-RNN discussed in~\eqref{eq:RNN}, let $z[n]$ be the sequence of matricized empirical moment tensors $\text{Mat}\left(\hat{\Ebb}[y_i \otimes \Pc_2(x[n], i)]\right)$ for $i \in [n]$. Then,
%\begin{align*}
% \| z - \Ebb(z) \|  &\leq G\frac{1+\frac{1}{\sqrt{8}cn^{1.5}}}{1-\theta}\sqrt{8c^2n\log\left(\frac{d_y+d_x^2}{\delta}\right)},
%\end{align*}
% with probability at least $1-\delta$, where $\Ebb(z)$ is expectation over samples of Markov chain when the initial distribution is the stationary distribution and $c$ is specified in Equation~\eqref{eqn:lipschitz-const} and proved above.
%
%\end{theorem}

%For proof see Appendix~\ref{sec:Appendix-concentration}.

\subsection{Proof of Theorem~\ref{thm:concentation}} \label{sec:Appendix-concentration}
In order to get the complete concentration bound in Theorem~\ref{thm:concentation}, we need Lemma~\ref{lemma:LipschitzRNNcomplete} in addition to the following Theorem.

 \begin{theorem}(Concentration bound for a matrix-valued function of a Markov chain] \label{thm:conc} 
 Consider a Markov chain with observation samples $x[n]=(x_1, \dotsc, x_n) \in S^n$, geometric ergodicity $G$, contraction coefficient $\theta$ and an arbitrary initial distribution. For any $c$-Lipschitz matrix-valued function $\Phi(\cdot): S^n \rightarrow \mathbb{R}^{d_1 \times d_2}$, we have
 \begin{align*}
 \Vert \Phi - \Ebb[\Phi] \Vert \leq G\frac{1+\frac{1}{\sqrt{8}cn^{1.5}}}{1-\theta}\sqrt{8c^2n\log\left(\frac{d_1+d_2}{\delta}\right)},
\end{align*}
 with probability at least $1-\delta$, where $\Ebb(\Phi)$ is expectation over samples of Markov chain when the initial distribution is the stationary distribution.
 \end{theorem}

\bprf
 The proof follows result of~\citep{kontorovich2008concentration},~\citep{konidaris2014hidden}, and Matrix Azuma theorem (which can be proved using the analysis of~\citep{tropp2012user} for sum of random matrices). 
 %For details, see Appendix H and I in~\citep{azizzadenesheli2016reinforcement}.
The upper bound can be decoupled into two parts, (1) $\Vert \Phi - \Ebb[\Phi] \Vert$ where the expectation is over the same initial distribution as used for $\Phi$ and (2) the difference between $\Ebb[\Phi]$ for the case where the initial distribution is the same initial distribution as used for $\Phi$ and the initial distribution being equal to the stationary distribution. It is direct from analysis of~\citep{kontorovich2014uniform} that the latter is upper bounded by $\sum_i G \theta^{-(i-1)} \leq \frac{G}{1-\theta}$. The former can be bounded by Theorem~\ref{thm:mAzuma} below and hence Theorem~\ref{thm:conc} follows.
\eprf
\begin{theorem}[Matrix Azuma~\citep{azizzadenesheli2016reinforcement}]\label{thm:mAzuma} 
Consider Hidden Markov Model with finite sequence of $n$ samples $S_i$ as observations given arbitrary initial states distribution and c-Lipschitz matrix valued function $\Phi: S_1^n \rightarrow \mathbb{R}^{d_1 \times d_2}$, then
\begin{align*}
\Vert \Phi - \Ebb[\Phi] \Vert \leq \frac{1}{1-\theta} \sqrt{8c^2n \log \left(\frac{d_1+d_2}{\delta}   \right)},
\end{align*}
with probability at least $1-\delta$. The $\Ebb[\Phi]$ is given the same initial distribution of samples.
\end{theorem}
\bprf
This proof is from \citep{azizzadenesheli2016reinforcement} and is repeated here for completeness. 
Theorem~7.1~\citep{tropp2012user} provides the upper confidence bound for summation of matrix random variables. Consider a finite sequence of matrices $\Psi_i \in \mathbb{R}^{d_1 \times d_2}$. The variance parameter $\sigma^2$ is the upper bound for $\sum_i \left[ \Psi_i - \Ebb_{i-1}[\Psi_i] \right], \forall i$ and we have that
\begin{align*}
\Vert \sum_i \left[ \Psi_i - \Ebb_{i-1}[\Psi_i] \right] \Vert \leq \sqrt{8\sigma^2 \log{\frac{d_1+d_2}{\delta}}},
\end{align*}
with probability at least $1-\delta$.
For function $\Phi$, we define the martingale difference of function $\Phi$ as the input random variable with arbitrary initial distribution over states.
\begin{align*}
\text{MD}_i(\Phi ; S_1^i) = \Ebb[\Phi | S_1^i] - \Ebb[\Phi | S_1^{i-1}],
\end{align*}
where $S_i^j$ is the subset of samples from i-th position in sequence to j-th one. Hence, the summation over these set of random variables gives $\Ebb[\Phi | S_1^n] - \Ebb[\Phi] = \Phi( S_1^n) - \Ebb[\Phi]$, $\Ebb[\Phi]$ is the expectation with the same initial state distribution. 

Then it remains to find $\sigma$ which is the upper bound for $\Vert \text{MD}_i(\Phi ; S_1^i)\Vert$ for all possible sequences. Define $\text{MD}_i(\Phi) = \underset{S_1^i}{\max}~{\text{MD}_i(\Phi ; S_1^i)}$. By~\citep{kontorovich2014uniform}, $\text{MD}_i(\Phi)$ is a c-Lipschitz  function and is upper bounded by $G\theta(n-i).$

\eprf
 
 %For details, see~\hscomment{cite}.
%\hscomment{add more details here from Kavita's paper. The proof is straightforward just write it down}
%\hscomment{Lipschitz is w.r.t\ Hamming metric and definition is w.r.t. l2 norm and we take the avg to find the bound. show the details s.t. this all makes sense}
% Using the Lipschitz constant derived in  in Theorem~\ref{thm:conc} completes the proof of Theorem~\ref{thm:concentation}.

Considering the analysis of tensor decomposition analysis in~\citep{JanzaminEtal:NN2015}, Theorem~\ref{thm:concentation} implies polynomial sample complexity for GLOREE. The sample complexity is a polynomial of $(d_x, d_y, d_h, G, \frac{1}{1-\theta}$, \\$ \sigma_{\min}^{-1}(A_1), \sigma_{\min}^{-1}(A_2), \sigma_{\min}^{-1}(U)). $ Detailed proof is similar to analysis in~\citep{JanzaminEtal:NN2015}, ~\citep{azizzadenesheli2016reinforcement}.
Note that with similar analysis we can prove polynomial sample complexity for GLOREE-B.

%\hscomment{add proof or more detailed analysis here} 
\subsection{Remark on Assumptions 2(d), 2(e)} \label{appendix:remark}

As you saw in the proof of lemma~\ref{lemma:LipschitzRNNcomplete}, in order to prove the Lipschitz property for moment tensor, we need the following:

\begin{align*}
\frac{1}{n}\sum_{i \in [n]} \|\nabla_{x_i} \left[y_t \otimes \Pc_2(x[n], t) \right]
=\frac{1}{n}\sum_{i \in [n]} \left \lbrace\| \Pc_2(x[n], i) \nabla_{x_i} y_t  + y_t \otimes \nabla_{x_i}\Pc_2(x[n], t)\| \right\rbrace =
 O(\frac{1}{n})
\end{align*}

By result of lemma~\ref{lemma:LipschitzRNN}, in order to prove the $O(1/n)$ for the first term we need $\Vert\Pc_2(x[n], i)\Vert$ to be a bounded value.

As we showed in Appendix~\ref{appendix:Liptensor}, Assumptions 2(d), 2(e) suffice to prove the bound for the second term too. However, that is not a necessary assumption. We provide an example here. 

\paragraph{Example: } Assumptions (2d)-(2e) can be replaced by the following:
\bi
\item[(2d*)] $ \nabla_{x_i}\Pc_2(x[n], t), i \in [n]$ is nonzero only for $\alpha$ terms and is zero otherwise, i.e., Let $\Omega = \lbrace i \in [n],  \nabla_{x_i}\Pc_2(x[n], t) \neq 0 \rbrace,$ we assume $ |\Omega | \leq \alpha $. For example, if input sequence is a first order Markov chain as in \eqref{eqn:sf-mc}, we have $\alpha = 3$, i.e., the term is nonzero only for $i = t-1, t, t+1$. In general, this holds for Markov chains of higher order $p$ such that $p << n$. 
\item[(2e*)] For those values of $i$ such that $ \nabla_{x_i}\Pc_2(x[n], t), i \in [n]$ is nonzero, it is bounded by some constant $\gamma$ which is specified by pdf of the input. \ei

It can be readily seen from the proof in Appendix~\ref{appendix:Liptensor} that if we replace Assumptions  2(d), 2(e), we can still prove the result. 
%\bi
%\item $ \nabla_{x_i}\Pc_2(x[n], t), i \in [n]$ is nonzero only for $\alpha$ terms and is zero otherwise, i.e., Let $\Omega = \lbrace i \in [n],  \nabla_{x_i}\Pc_2(x[n], t) \neq 0 \rbrace,$ we assume $ |\Omega | \leq \alpha $. For example, if input sequence is a first order Markov chain as in \eqref{eqn:sf-mc}, we have $\alpha = 3$, i.e., the term is nonzero only for $i = t-1, t, t+1$. %\hscomment{can we come up with a more general/understandable case?}
%\item For those values of $i$ such that $ \nabla_{x_i}\Pc_2(x[n], t), i \in [n]$ is nonzero, it is bounded by some constant $\gamma$ which is specified by pdf of the input. \ei

In this case, the $3\gamma$ in bound for $c$ will be replaced by $\alpha \gamma$. Note that first order Markov chain is a special case of this example.

\section{Discussion}

\subsection{Score Function Estimation} \label{sec:sf-estimation}
According to~\citep{janzamin2014matrix}, there are various efficient methods for estimating the score function. The framework of score matching is popular for parameter estimation  in probabilistic models~\citep{hyvarinen2005estimation, swersky2011autoencoders}, where the criterion is to fit parameters based on matching the data score function. \citet{swersky2011autoencoders} analyze the score matching for latent energy-based models.
In deep learning, the framework of auto-encoders attempts to find encoding and decoding functions which minimize the reconstruction error under added noise; the so-called Denoising Auto-Encoders (DAE). This is an unsupervised framework involving only unlabeled samples. \citet{alain2012regularized} argue that the DAE   approximately learns the first order score function of the input, as the noise variance goes to zero. ~\citet{sriperumbudur2013density} propose non-parametric score matching methods that provides the non-parametric score function form for infinite dimensional exponential families with guaranteed convergence rates. Therefore, we can use any of these methods for estimating $\Pc_1(x[n], [n])$ and use the recursive form~\citep{janzamin2014matrix}.
%in~\eqref{eqn:diffoperator_recursion_informal} to estimate higher order score function.
%
%. The framework of score matching is popular for parameter estimation  in probabilistic models~\citep{hyvarinen2005estimation, swersky2011autoencoders}, where the criterion is to fit parameters based on matching the data score function. As another method, the framework of denoising auto-encoder (DAE) is studied by \citet{alain2012regularized} to argue that DAE   approximately learns the first order score function of the input. Thus, we can use any of these methods for estimating $\Pc_1(x)$ and use the recursive form
$$\Pc_m(x[n], [n]) = - \Pc_{m-1}(x[n], [n]) \otimes \nabla_{x[n]} \log p(x[n]) - \nabla_{x[n]} \Pc_{m-1}(x[n],[n])$$ to estimate higher order score functions.

\subsection{Training IO-RNN and BRNN with scalar output} \label{Appendix:scalar}
In the main text, we discussed training IO-RNNs and BRNNs with vector outputs. Here we expand the results to training IO-RNNs and BRNNs with scalar outputs. Note that in order to recover the parameters uniquely, we need the cross-moment to be a tensor of order at least $3$. This is due to the fact that in general matrix decomposition does not provide unique decomposition for non-orthogonal components. In order to obtain a cross-moment tensor of order at least $3$, since the output is scalar, we needs its derivative tensors of order at least $3$. In order to have a non-vanishing gradient,  the activation function needs to be a polynomial of order $l \geq 3$.

Hence,  our method can also be used for training IO-RNN  and BRNN with scalar output if the activation function is a polynomial of order $l \geq 3$, i.e.,
Let $y_t$ be the output of 
\begin{align*}
\Ebb[ y_t| h_t] = \inner{a_2,h_t}, \quad
h_t = \poly_l(A_1 x_{t}+ U h_{t-1}),
\end{align*}
where
$x_t \in \mathbb{R}^{d_x}, h_t \in \mathbb{R}^{d_h}, y_t \in \mathbb{R}$ and hence $A_1 \in \mathbb{R}^{d_h \times d_x}, U \in \mathbb{R}^{d_h \times d_h}$, $a_2 \in \mathbb{R}^{d_h}$. We can learn the parameters of the model using GLOREE with guarantees. 

We have
\begin{lemma}[Learning parameters of RNN for general activation function, scalar output]
\begin{align*}
\Ebb \left[ y_t \otimes \Sc_3(x_t) \right] = \sum_{i \in d_h} \mu_i {a_2}_i  A_1^{(i)} \otimes A_1^{(i)}  \otimes A_1^{(i)},\\
\mu_i = \Ebb \left[ \left( \inner{A_1^{(i)}, x_t}+\inner{U^i,h_{t-1}}     \right)^{*(l-3)}     \right]
\end{align*}
In order to learn $U$, we form the tensor $\hat{\Ebb} \left[y_t \otimes \text{Reshape}(\Sc_{l^2}(x_{t-1}),1, [1 ~\dotsc~ l],\dotsc, [l^2-l+1~ \dotsc ~l^2] \right]$.
Then we have
\begin{align*}
\Ebb \left[ y_t\otimes \text{Reshape}(\Sc_{l^2}(x_{t-1}),1, [1 ~\dotsc~ l],\dotsc, [l^2-l+1~ \dotsc ~l^2] \right]  =  \sum_{i \in d_h} (a_2)_i   \otimes  \left[\left[U (A_1^{\odot l}) \right]^{(i)}\right]^{\otimes l},
\end{align*}
where $\odot$ is the row-wise Kronecker product defined in~\ref{KR}.
Hence, since we know $A_1$, we can recover $U$ via tensor decomposition.
\end{lemma}

We can prove that we can learn the parameters of a BRNN with scalar output and polynomial activation functions of order $l \geq 3$ using the same trend as for Lemma~\ref{lemma:BRNN}.

%\hscomment{add more detail/expalanation here}

\subsection{Training Linear IO-RNN} \label{sec:linear}
In the paper we discussed the problem of training IO-RNNs with polynomial activation function of order $l \geq 2$. Here we propose a method for training IO-RNNs with linear activation functions. Although our proposed methods for two cases differ in nature, we include both of them for completeness and covering all cases.

% a different approach for linear activation
\citet{Sedghi:SparseNet} provide a method to train first layer of feed-forward neural networks using the first-order score function of the input. For a NN with vector output, their formed cross-moment is a matrix of the form $\Ebb[ y \otimes \Pc_1(x)]= BA_1,$ where $A_1$ is the weight matrix for the first layer and $B$ is a matrix that includes the rest of the derivative matrix. Then they argie that if $A_1$ is sparse, the problem of recovering $A_1$ is a sparse dictionary learning problem that can be solved efficiently using Sparse Dictionary Learning Algorithm~\citep{Spielman-12}.

Here we show that for IO-RNN with linear activation function, we can expand the result of~\citet{Sedghi:SparseNet} to the non-i.i.d.\ input sequence. 

Let
\begin{align*}
y_t = A_2^\top h_t, \quad
h_t = A_1 x_{t}+ U h_{t-1},
\end{align*}
where $x_t \in \mathbb{R}^{d_x}, h_t \in \mathbb{R}^{d_h}, y_t \in \mathbb{R}^{d_y}, A_2^\top \in \mathbb{R}^{d_h \times d_y}$ and hence $A_1 \in \mathbb{R}^{d_y \times d_x}, U \in \mathbb{R}^{d_y \times d_h}$.

%Let $\tl{y}[n]=[y_1, y_2,\dotsc, y_n], \tl{x}[n]=[x_1, x_2,\dotsc, x_n]$. Similar to our earlier analysis, we have\begin{align*}
%\Ebb[ \tl{y}[n] \otimes \Pc(\tl{x}[n],[n]) =\nabla_{\tl{x[n]}} \tl{y}[n] 
%\end{align*}
%For our linear model the derivative has a Toeplitz form. Assuming that $A_1$ is sparse, we can use this structure and Sparse Dictionary Learning Algorithm~\citep{Spielman-12} to recover the model parameters.

Let $\tl{y}[n]=[y_1, y_2,\dotsc, y_n], \tl{x}[n]=[x_1, x_2,\dotsc, x_n]$. Similar to our earlier analysis, we have\begin{align*}
\Ebb[ \tl{y}[n] \otimes \Pc(\tl{x}[n],[n]) =\nabla_{\tl{x}[n]} \tl{y}[n]
\end{align*}
For our linear model the derivative has a Toeplitz form. Assuming that $A_1$ is sparse, we can use this structure and Sparse Dictionary Learning Algorithm~\citep{Spielman-12} to recover the model parameters.

Below we write the cross-moment Topelitz form for $n=4$ for simplicity.
\begin{align*}
 \Ebb[ \tl{y}[n] \otimes \Pc(\tl{x}[n],[n]) =\left[ \begin{array}{cccc} A_2^\top A_1 & 0 & 0 & 0\\
A_2^\top U A_1 & A_2^\top A_1 & 0 & 0\\
A_2^\top U^2 A_1 & A_2^\top U A_1 &  A_2^\top A_1 &  0\\
A_2^\top U^3 A_1& A_2^\top U^2 A_1 & A_2^\top U A_1 & A_2^\top A_1\\
\end{array} \right]
\end{align*}
%In order to recover the parameters, from the above, we note that $\nabla_{x_i} y_j = A_2^\top U^{j-i} A_1$. Inspired by this structure, we form the matrix
If we recover the Toeplitz structure, we have access to the following matrices: $A_2 A_1, A_2 U A_1, \dotsc, A_2 U ^n A_1$. Next, we put these matrices in a new matrix $C$ as below.
\begin{align*}
C=\left[\begin{array}{c}
A_2 A_1\\
A_2 U  A_1\\
\vdots\\
A_2  U^n A_1
\end{array}\right]
\end{align*}
It is easy to see that $C= BA_1$ for  matrix $B$ as shown below
\begin{align*}
B=\left[\begin{array}{c}
A_2 A_1\\
A_2 U  A_1\\
\vdots\\
A_2  U^n A_1
\end{array}\right]
\end{align*}
Now assuming that $A_1$ is sparse and $B$ is full column-rank, we can recover $A_1$ using Sparse Dictionary Learning Algorithm~\citep{Spielman-12}.

Let $U=V \Lambda V^\top$ be the singular-value decomposition of $U$, where $\Lambda=\Diag(\lambda_i)$. It is easy to show that, to ensure that  $B$ is full column-rank, we need the singular-values of $U$ to satisfy $\lambda_i \sim \frac{1}{\sqrt{d_h}}$. Once we recover $A_1$, we can recover $A_2 = A_2 A_1 A_1^{-1}$ and $U=A_2^{-1} A_2 U  A_1 A_1^{-1}$.

\section{Spectral Decomposition Algorithm} \label{sec:tensor}

As part of GLOREE, we need a spectral/tensor method to decompose the cross-moment tensor to its rank-1 components. Refer to notation for definition of tensor rank and its rank-1 components. As depicted in notation, we are considering CP decomposition.  Note that CP tensor decomposition for various scenarios is extensively analyzed in the literature~\citep{SongEtal:NonparametricTensorDecomp},~\citep{AnandkumarEtal:tensor12},~\citep{ JanzaminEtal:Altmin14},~\citep{anandkumar2014guaranteed},~\citep{janzamin2014matrix},~\citep{ JanzaminEtal:NN2015}. We follow the method in~\citep{JanzaminEtal:NN2015}.

\begin{figure}[t]
%\begin{wrapfigure}{r}{0.3\textwidth}
%\vspace{-.35in}
\bc
%\resizebox{0.4\textwidth}{!}{
\begin{tikzpicture}
[
scale=1,
   nodestyle/.style={fill = gray!30, shape = rectangle, rounded corners, minimum width = 2cm},
]
\small
\matrix [column sep=2mm,row sep=3mm] {
\node[nodestyle](a1){Input: Tensor $T=\sum_{i \in [k]} \lambda_i u_i^{\otimes 3}$}; & \\
\node[nodestyle](w){Whitening procedure}; & \\
\node[nodestyle](a){SVD-based Initialization }; & \\
\node[nodestyle, align=center](b){Tensor Power Method}; & \\
%\node[nodestyle, align=center](c){Clustering the output of \\ tensor power method \\ into $k$ clusters}; &
%\node[nodestyle, align=center](c2){Labeled data: \\ $\{(x_i,y_i)\}$};
%\node[nodestyle, align=center](d){Removing the residual error }; & \\
\node[nodestyle](e){Output: $\lbrace u_i  \rbrace_{i \in [k]}$}; & \\
%\node[nodestyle, align=center](e){Spectral/tensor method: \\ find $u_j$'s s.t.\ $\displaystyle\Ebb \left[ \nabla^{(m)} G(x) \right]= \sum_{j \in [k]} u_j^{\otimes m}$}; & \\
%\node[nodestyle, align=center](f){Extract discriminative features using $u_j$'s/ \\ do model-based prediction with $u_j$'s as parameters}; & \\
};
\draw [->, line width = 1pt] (a1) to (w);
\draw [->, line width = 1pt] (w) to (a);
\draw [->, line width = 1pt] (a) to (b);
\draw [->, line width = 1pt] (b) to (e);

\end{tikzpicture}
%}
\ec
%\vspace{-7pt}
\caption{\small Overview of tensor decomposition algorithm for a symmetric third order tensor~\citep{JanzaminEtal:NN2015}.}% The details of the specified algorithms are provided in the Appendix.}
\label{fig:TensorDecomposition}
\end{figure}
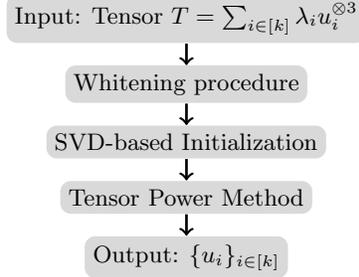

The only difference between our tensor decomposition setting and that of~\citep{JanzaminEtal:NN2015} is that they have a symmetric tensor (i.e., $\hat{T}=\sum_{i \in [r]} c_i \otimes c_i \otimes c_i$) whereas 
in GLOREE, we have two asymmetric tensor decomposition procedures in the form of $\hat{T} = \sum_{i \in [r]} b_i \otimes c_i \otimes c_i $. 
Therefore, \bi
\item[1] We first make a symmetric version of our tensor. For our specific case, 
this includes multiplying the first mode of the tensor with a matrix $D$, such that  $ \hat{T}(D,I,I) \simeq \sum_{i \in [r]} c_i \otimes c_i \otimes c_i$. We use the rule presented in~\citep{anandkumar2012spectral} to form the symmetrization tensor.  For example for $\hat{T} = \Ebb[ y_i \otimes \mathcal{S}_2(x[n], i)]$, we use $\hat{D} = [\Ebb[ y_i \otimes \mathcal{S}_1(x[n], i)]]^{-1}$.
\item[2] Next, we run the tensor decomposition procedure as in~\citep{JanzaminEtal:NN2015} to recover estimates of $\hat{c}_i, i \in [r]$. The steps are shown in Figure~\ref{fig:TensorDecomposition}. For more details, see~\citep{JanzaminEtal:NN2015}. 
\item[3] The last step includes reversing the effect of symmetrization matrix $D$ to recover estimate of $\hat{b}_i, i \in [r]$. For more discussion on symmetrization, see~\citep{anandkumar2012spectral}. 
\end{itemize}

Our overall algorithm is shown in Algorithm~\ref{algo:main}.
 
\paragraph{Remark on tensor decomposition via sketching: } Consider line~\ref{line:sketching} in Algorithm~\ref{algo:mainRNN_poly} and line~\ref{line:Bsketching} in Algorithm~\ref{algo:BRNN-general}. Here we are decomposing a tensor of order $l+1$. The tensor decomposition algorithm for third order tensor readily generalizes to higher order tensors. In order to perform this with efficient computational complexity, we can use tensor sketching proposed by~\citet{wang2015fast}. They do not form the moment tensor explicitly and directly compute tensor sketches from data. This avoids the exponential blowup in computation, i.e., it reduces the computational complexity from $m^{l+1}$ to $(m+m\log m)n$, where $m$ is the sketch length and $n$ denotes the number of samples. As expected, there is a trade off between the sketch length and the error in recovering the tensor components. For details, see~\citep{wang2015fast}.
%After the unwhitening procedure we reverse the effect of symmetrization.
%For a more compact representation, we use the same tensor decomposition algorithm for both cases. Note that the symmetric case can be considered as a specific form of asymmetric case where the symmetrization matrix is identity.  
% The second tensor decomposition we need to do is on the reshaped tensor $$\left[y_t\otimes \text{Reshape}(\Sc_{l^2}(x[n],{t-1}),1, [1 ~\dotsc~ l],\dotsc, [l^2-l+1~ \dotsc ~l^2] \right].$$ Note that this is a tensor of order $l+1$, where the modes $2, \dotsc, l^2$ are similiar. In order to reduce computational complexity, we contract the tensor such that we get a symmetric-third order tensor. 
%The rest of the tensor decomposition procedure is the same as~\citep{JanzaminEtal:NN2015}. For more details, see~\citep{JanzaminEtal:NN2015}. 
%The steps we take for tensor decomposition are shown in Figure~\ref{fig:TensorDecomposition} and the Algorithm~\ref{algo:main} provides the details. 

%\vspace{-.4in}
%\end{wrapfigure}

%
\floatname{algorithm}{Algorithm}
\begin{algorithm}[t]
\caption{Tensor Decomposition Algorithm Setup}
\label{algo:main}
\begin{algorithmic}[1]
\renewcommand{\algorithmicrequire}{\textbf{input}}
\renewcommand{\algorithmicensure}{\textbf{output}}
\REQUIRE Asymmetric tensor $T$, symmetrization matrix $D$.

%\STATE Compute $\widehat{M}_3=\frac{1}{n}\sum_i y_i \Pc_3(x_i)$, Empirical estimate of $M_3$. % $\widehat{M}_3=\frac{1}{n}\sum_n y_i \cdot x_i^{\otimes 3}-\widetilde{R}$.
%\IF{Whitening}
\STATE Symmetrize the tensor : $T = T(D, I, I)$.
%\STATE Calculate $T=$Whiten($T$). %; see Procedure~\ref{algo:whitening}.
%\ELSIF{Tensorizing}
%\STATE Tensorize the input tensor.
%\STATE Calculate $T=$Whiten($T$); see Procedure~\ref{algo:whitening}.
%\ENDIF
%\STATE $\lbrace u_j , \lambda_j \rbrace_{j \in [k]}=\text{tensor power decomposition}(T)$. (Algorithm~\ref{alg:robustpower} in the Appendix)
%\FOR{$j=1$ to $k$}
%\STATE $(v_j , \mu_j, T)=\text{tensor power decomposition}(T)$ %; see Algorithm~\ref{alg:robustpower}.
%\ENDFOR
%\STATE Compute $u_1$ and $v_1$ as the top left and right singular vectors of  $T(I,I,\theta) \in \R^{d \times d}$.
%\STATE $\ha^{(0)} \leftarrow u_1$, $\hb^{(0)} \leftarrow v_1$.
%\STATE Initialize $\hc^{(0)}$ by update formula in \eqref{eqn:asymmetric power update}.
\STATE $(A_1)_j = \operatorname{Tensor Decomposition}(T)$ as in Figure~\ref{fig:TensorDecomposition}. For details, see~\citep{JanzaminEtal:NN2015}.
%; see Procedure~\ref{algo:Un-whitening}.
\STATE $A_2 =D^{-1}A_1$
\RETURN $A_2, A_1, A_1.$
\end{algorithmic}
\end{algorithm}


\begin{thebibliography}{30}
\providecommand{\natexlab}[1]{#1}
\providecommand{\url}[1]{\texttt{#1}}
\expandafter\ifx\csname urlstyle\endcsname\relax
  \providecommand{\doi}[1]{doi: #1}\else
  \providecommand{\doi}{doi: \begingroup \urlstyle{rm}\Url}\fi

\bibitem[Alain and Bengio(2012)]{alain2012regularized}
Guillaume Alain and Yoshua Bengio.
\newblock What regularized auto-encoders learn from the data generating
  distribution.
\newblock \emph{arXiv preprint arXiv:1211.4246}, 2012.

\bibitem[Anandkumar et~al.(2014{\natexlab{a}})Anandkumar, Ge, Hsu, Kakade, and
  Telgarsky]{AnandkumarEtal:tensor12}
A.~Anandkumar, R.~Ge, D.~Hsu, S.~M. Kakade, and M.~Telgarsky.
\newblock Tensor decompositions for learning latent variable models.
\newblock \emph{J. of Machine Learning Research}, 15:\penalty0 2773--2832,
  2014{\natexlab{a}}.

\bibitem[Anandkumar et~al.(2012)Anandkumar, Liu, Hsu, Foster, and
  Kakade]{anandkumar2012spectral}
Anima Anandkumar, Yi-kai Liu, Daniel~J Hsu, Dean~P Foster, and Sham~M Kakade.
\newblock A spectral algorithm for latent dirichlet allocation.
\newblock In \emph{Advances in Neural Information Processing Systems}, pages
  917--925, 2012.

\bibitem[Anandkumar et~al.(2014{\natexlab{b}})Anandkumar, Ge, Hsu, Kakade, and
  Telgarsky]{JMLR:v15:anandkumar14b}
Anima Anandkumar, Rong Ge, Daniel Hsu, Sham~M. Kakade, and Matus Telgarsky.
\newblock Tensor decompositions for learning latent variable models.
\newblock \emph{Journal of Machine Learning Research}, 15:\penalty0 2773--2832,
  2014{\natexlab{b}}.

\bibitem[Anandkumar et~al.(2014{\natexlab{c}})Anandkumar, Ge, and
  Janzamin]{JanzaminEtal:Altmin14}
Anima Anandkumar, Rong Ge, and Majid Janzamin.
\newblock {Sample Complexity Analysis for Learning Overcomplete Latent Variable
  Models through Tensor Methods}.
\newblock \emph{arXiv preprint arXiv:1408.0553}, Aug. 2014{\natexlab{c}}.

\bibitem[Anandkumar et~al.(2014{\natexlab{d}})Anandkumar, Ge, and
  Janzamin]{anandkumar2014guaranteed}
Anima Anandkumar, Rong Ge, and Majid Janzamin.
\newblock {Guaranteed Non-Orthogonal Tensor Decomposition via Alternating
  Rank-1 Updates}.
\newblock \emph{arXiv preprint arXiv:1402.5180}, Feb. 2014{\natexlab{d}}.

\bibitem[Azizzadenesheli et~al.(2016)Azizzadenesheli, Lazaric, and
  Anandkumar]{azizzadenesheli2016reinforcement}
Kamyar Azizzadenesheli, Alessandro Lazaric, and Anima Anandkumar.
\newblock Reinforcement learning of pomdp's using spectral methods.
\newblock \emph{arXiv preprint arXiv:1602.07764}, 2016.

\bibitem[Balduzzi and Ghifari(2016)]{BalduzziStrong}
David Balduzzi and Muhammad Ghifari.
\newblock Strongly-typed recurrent neural networks.
\newblock \emph{arXiv preprint arXiv:1602.02218}, 2016.

\bibitem[Ben-Hur and Brutlag(2006)]{ben2006sequence}
Asa Ben-Hur and Douglas Brutlag.
\newblock Sequence motifs: highly predictive features of protein function.
\newblock In \emph{Feature Extraction}, pages 625--645. Springer, 2006.

\bibitem[Chen and Manning(2014)]{chen2014fast}
D.~Chen and C.~Manning.
\newblock A fast and accurate dependency parser using neural networks.
\newblock In \emph{EMNLP}, pages 740--750, 2014.

\bibitem[Hammer(2000)]{hammer2000approximation}
Barbara Hammer.
\newblock On the approximation capability of recurrent neural networks.
\newblock \emph{Neurocomputing}, 31\penalty0 (1):\penalty0 107--123, 2000.

\bibitem[Huang et~al.(2014)Huang, Ge, Kakade, and Dahleh]{huang2014minimal}
Qingqing Huang, Rong Ge, Sham Kakade, and Munther Dahleh.
\newblock Minimal realization problem for hidden markov models.
\newblock In \emph{Communication, Control, and Computing (Allerton), 2014 52nd
  Annual Allerton Conference on}, pages 4--11. IEEE, 2014.

\bibitem[Hyv{\"a}rinen(2005)]{hyvarinen2005estimation}
Aapo Hyv{\"a}rinen.
\newblock Estimation of non-normalized statistical models by score matching.
\newblock In \emph{Journal of Machine Learning Research}, pages 695--709, 2005.

\bibitem[Janzamin et~al.(2015)Janzamin, Sedghi, and
  Anandkumar]{JanzaminEtal:NN2015}
M.~Janzamin, H.~Sedghi, and A.~Anandkumar.
\newblock {Beating the Perils of Non-Convexity: Guaranteed Training of Neural
  Networks using Tensor Methods}.
\newblock \emph{Preprint available on arXiv:1506.08473}, June 2015.

\bibitem[Janzamin et~al.(2014)Janzamin, Sedghi, and
  Anandkumar]{janzamin2014matrix}
Majid Janzamin, Hanie Sedghi, and Anima Anandkumar.
\newblock {Score Function Features for Discriminative Learning: Matrix and
  Tensor Frameworks}.
\newblock \emph{arXiv preprint arXiv:1412.2863}, Dec. 2014.

\bibitem[Karpathy et~al.(2014)Karpathy, Toderici, Shetty, Leung, Sukthankar,
  and Fei-Fei]{karpathy2014large}
Andrej Karpathy, George Toderici, Sanketh Shetty, Thomas Leung, Rahul
  Sukthankar, and Li~Fei-Fei.
\newblock Large-scale video classification with convolutional neural networks.
\newblock In \emph{Proceedings of the IEEE conference on Computer Vision and
  Pattern Recognition}, pages 1725--1732, 2014.

\bibitem[Konidaris and Doshi-Velez(2014)]{konidaris2014hidden}
George Konidaris and Finale Doshi-Velez.
\newblock Hidden parameter markov decision processes: An emerging paradigm for
  modeling families of related tasks.
\newblock In \emph{2014 AAAI Fall Symposium Series}, 2014.

\bibitem[Kontorovich and Weiss(2014)]{kontorovich2014uniform}
Aryeh Kontorovich and Roi Weiss.
\newblock Uniform chernoff and dvoretzky-kiefer-wolfowitz-type inequalities for
  markov chains and related processes.
\newblock \emph{Journal of Applied Probability}, 51\penalty0 (4):\penalty0
  1100--1113, 2014.

\bibitem[Kontorovich et~al.(2008)Kontorovich, Ramanan,
  et~al.]{kontorovich2008concentration}
Leonid~Aryeh Kontorovich, Kavita Ramanan, et~al.
\newblock Concentration inequalities for dependent random variables via the
  martingale method.
\newblock \emph{The Annals of Probability}, 36\penalty0 (6):\penalty0
  2126--2158, 2008.

\bibitem[Lipton et~al.(2015)Lipton, Berkowitz, and Elkan]{lipton2015critical}
Zachary~C Lipton, John Berkowitz, and Charles Elkan.
\newblock A critical review of recurrent neural networks for sequence learning.
\newblock \emph{arXiv preprint arXiv:1506.00019}, 2015.

\bibitem[Manning and Sch{\"u}tze(1999)]{manning1999foundations}
C.~Manning and H.~Sch{\"u}tze.
\newblock \emph{Foundations of statistical natural language processing}, volume
  999.
\newblock MIT Press, 1999.

\bibitem[Schuster and Paliwal(1997)]{schuster1997bidirectional}
Mike Schuster and Kuldip~K Paliwal.
\newblock Bidirectional recurrent neural networks.
\newblock \emph{Signal Processing, IEEE Transactions on}, 45\penalty0
  (11):\penalty0 2673--2681, 1997.

\bibitem[Sedghi and Anandkumar(2014)]{Sedghi:SparseNet}
Hanie Sedghi and Anima Anandkumar.
\newblock Provable methods for training neural networks with sparse
  connectivity.
\newblock \emph{NIPS workshop on Deep Learning and Representation Learning},
  Dec. 2014.

\bibitem[Song et~al.(2013)Song, Anandkumar, Dai, and
  Xie]{SongEtal:NonparametricTensorDecomp}
L.~Song, A.~Anandkumar, B.~Dai, and B.~Xie.
\newblock Nonparametric estimation of multi-view latent variable models.
\newblock \emph{arXiv preprint arXiv:1311.3287}, Nov. 2013.

\bibitem[Spielman et~al.(2012)Spielman, Wang, and Wright]{Spielman-12}
D.~Spielman, H.~Wang, and J.~Wright.
\newblock Exact recovery of sparsely-used dictionaries.
\newblock In \emph{Conference on Learning Theory}, 2012.

\bibitem[Sriperumbudur et~al.(2013)Sriperumbudur, Fukumizu, Kumar, Gretton, and
  Hyv{\"a}rinen]{sriperumbudur2013density}
Bharath Sriperumbudur, Kenji Fukumizu, Revant Kumar, Arthur Gretton, and Aapo
  Hyv{\"a}rinen.
\newblock Density estimation in infinite dimensional exponential families.
\newblock \emph{arXiv preprint arXiv:1312.3516}, 2013.

\bibitem[Swersky et~al.(2011)Swersky, Buchman, Freitas, Marlin,
  et~al.]{swersky2011autoencoders}
Kevin Swersky, David Buchman, Nando~D Freitas, Benjamin~M Marlin, et~al.
\newblock On autoencoders and score matching for energy based models.
\newblock In \emph{Proceedings of the 28th International Conference on Machine
  Learning (ICML-11)}, pages 1201--1208, 2011.

\bibitem[Tropp(2012)]{tropp2012user}
J.~Tropp.
\newblock User-friendly tail bounds for sums of random matrices.
\newblock \emph{Foundations of Computational Mathematics}, 12\penalty0
  (4):\penalty0 389--434, 2012.

\bibitem[Wang et~al.(2015)Wang, Tung, Smola, and Anandkumar]{wang2015fast}
Yining Wang, Hsiao-Yu Tung, Alexander Smola, and Anima Anandkumar.
\newblock Fast and guaranteed tensor decomposition via sketching.
\newblock In \emph{Proc. of NIPS}, 2015.

\bibitem[Xie et~al.(2015)Xie, Sun, Zhu, Chen, and Yu]{xie2015recurrent}
Qizhe Xie, Kai Sun, Su~Zhu, Lu~Chen, and Kai Yu.
\newblock Recurrent polynomial network for dialogue state tracking with
  mismatched semantic parsers.
\newblock In \emph{16th Annual Meeting of the Special Interest Group on
  Discourse and Dialogue}, page 295, 2015.

\end{thebibliography}
\end{document}